\documentclass[pmlr]{jmlr}

\usepackage{longtable}

\usepackage{booktabs}
\usepackage{multirow}
\usepackage{amsmath}
\DeclareMathOperator*{\argmax}{arg\,max}
\DeclareMathOperator*{\argmin}{arg\,min}
\usepackage{mathbbol}
 
\usepackage[load-configurations=version-1]{siunitx} 

\makeatletter
\def\set@curr@file#1{\def\@curr@file{#1}} 
\makeatother

\theorembodyfont{\upshape}
\theoremheaderfont{\scshape}
\theorempostheader{:}
\theoremsep{\newline}

\jmlrvolume{219}
\jmlryear{2023}
\jmlrworkshop{Machine Learning for Healthcare}

\title[Deep Metric Learning for Hemodynamics Inference with ECG]{Deep Metric Learning for the Hemodynamics Inference with Electrocardiogram Signals}

\author{\Name{Hyewon Jeong}
       \Email{hyewonj@mit.edu}\\ 
       \Name{Collin M. Stultz}
       \Email{cmstultz@mit.edu}\\ 
       \Name{Marzyeh Ghassemi}
       \Email{mghassem@mit.edu}\\ 
       \addr MIT CSAIL
       }

\begin{document}
\maketitle
\begin{abstract}
Heart failure is a debilitating condition that affects millions of people worldwide and has a significant impact on their quality of life and mortality rates. An objective assessment of cardiac pressures remains an important method for the diagnosis and treatment prognostication for patients with heart failure. Although cardiac catheterization is the gold standard for estimating central hemodynamic pressures, it is an invasive procedure that carries inherent risks, making it a potentially dangerous procedure for some patients. Approaches that leverage non-invasive signals – such as the electrocardiogram (ECG) – have the promise to make the routine estimation of cardiac pressures feasible in both inpatient and outpatient settings. Prior models trained to estimate intracardiac pressures (e.g., mean pulmonary capillary wedge pressure (mPCWP)) in a supervised fashion have shown good discriminatory ability but have been limited to the labeled dataset from the heart failure cohort. Furthermore, obtaining large datasets for diverse patient cohorts with intracardiac pressure labels is challenging due to the invasive nature of the procedure. To address these issues and build a robust representation, we apply traditional deep metric learning (DML) and propose a novel self-supervised DML with distance-based mining that improves the performance of a model with limited labels. We use a dataset that contains over 5.4 million ECGs without concomitant central pressure labels to pre-train a self-supervised DML model which showed improved classification of elevated mPCWP compared to self-supervised contrastive baselines. Additionally, the supervised DML model that uses ECGs with access to 8,172 mPCWP labels demonstrated significantly better performance on the mPCWP regression task compared to the supervised baseline. Moreover, our data suggest that DML yields models that are performant across patient subgroups, even when some patient subgroups are under-represented in the dataset. 
\end{abstract}
\section{Introduction}

Heart failure is a chronic condition that occurs when the heart is unable to pump blood effectively, which affects people worldwide with significant morbidity and mortality rates \citep{roger2013epidemiology}. The diagnosis and management of heart failure are challenging due to the heterogeneity in the phenotype of patients \citep{heidenreich20222022, roger2013epidemiology, burkhoff2003heart, komajda2014heart}. Accurately measuring intracardiac pressure is crucial for diagnosing and following up on treatment response in heart failure patients, as hemodynamic assessment can predict the risk of various negative outcomes \citep{baudry2022prognosis}. Intracardiac pressures play an important role in the assessment of hemodynamic severity in patients with heart failure \citep{heidenreich20222022}. The mean pulmonary capillary wedge pressure (mPCWP) - an estimate of the left atrial pressure - is one intracardiac measurement that is used for treatment and prognostication in heart failure patients \citep{drazner2008value, heidenreich20222022}. Right heart catheterization is a procedure that is used to measure intracardiac and pulmonary pressures (e.g., mPCWP). However, it is an invasive procedure, where a pressure transducer is inserted into a great vessel and threaded into the right heart chambers \citep{drazner2008value, heidenreich20222022}. To provide readily available diagnostic tools to detect and understand heart failure, we need safe routine screening methods that reliably predict patient hemodynamics.

One widely available clinical datum is the electrocardiogram (ECG), which has been applied for detecting arrhythmia \citep{hannun2019cardiologist}, myocardial infarction \citep{acharya2017application}, and heart failure \citep{masetic2016congestive, acharya2019deep}. Prior works have targeted supervised models for non-invasive estimation of hemodynamics using ECGs \citep{schlesinger2022deep, raghu2023ecg}. However, the population of catheterized patients with labeled data is limited, and supervised models do not leverage the rich information contained in large, unlabeled ECG datasets. Self-supervised models pre-trained on unlabeled ECGs such as PCLR~\citep{diamant2022patient}, CLOCS~\citep{kiyasseh2021clocs}, and SimCLR \citep{chen2020simple} have performed well in predicting cardiac abnormalities. However, our experiments found that contrastive learning methods perform worse than supervised models in regression tasks (Table \ref{result}). Further, we observed that supervised and contrastive baselines fail to achieve fair prediction across gender groups (Table \ref{subgroup}).

In this work, we seek to improve models by leveraging Deep Metric Learning (DML) with mining methods using either label or similarity distance metrics. DML is similarity-based learning where the objective is formulated to employ formal distance metrics between encoded representations to capture semantic similarities between data. We apply DML to a real-world clinical task, creating a supervised DML baseline where positive instances are sampled by label. We further propose a novel self-supervised DML model for time series and signal datasets where positive instances are sampled by a self-supervised distance-based ranking (Figure \ref{concept}). We also present a continuous label-based mining method for supervised DML. We focus on intracardiac pressure classification and regression using 12-lead ECGs, with 8,172 ECGs with matched mPCWP (\textit{ECGs with Labels}) and 5,426,614 unlabeled ECGs (\textit{ECGs without Labels}) from the data warehouse of Massachusetts General Hospital and Brigham and Women’s Hospital. We compare the DML models to state-of-the-art supervised and self-supervised models \citep{chen2020simple, kiyasseh2021clocs, diamant2022patient}, and evaluate their ability to predict intracardiac pressure values and classes (e.g., elevated mPCWP).

We found that supervised DML showed significant improvements in the regression of hemodynamics while obtaining competitive performance in classification, compared to the supervised and contrastive learning baselines (Table \ref{result}). Self-supervised DML showed significant improvements in the discriminability of the model with the regression performance compared to random initialization and the performance was on par with baselines, which shows promise that the utilization of unlabeled datasets helps achieve more robustness (Table \ref{result}). Additionally, we found that both supervised baseline and contrastive learning showed high subgroup performance gaps, while supervised DML and self-supervised DML significantly reduced the subgroup gaps (Table \ref{subgroup}, \ref{agegap}, Figure \ref{perf_gap}).

\subsection*{Generalizable Insights about Machine Learning in the Context of Healthcare}

Invasive procedures such as right cardiac catheterization result in limited data availability. Contrastive learning has been used to address the problem of restricted labels, but in our dataset, it resulted in poor regression performance and inequitable prediction across gender groups (Table \ref{result}). To improve inference in the presence of a limited label, we propose using both supervised and self-supervised DML in intracardiac pressure inference tasks (classification and regression). The supervised DML and self-supervised DML models outperformed the supervised and contrastive learning baselines in elevated mPCWP classification and regression tasks and achieved fairer predictions across different gender groups. Overall, our findings suggest that DML approaches can be utilized to create more robust representations for hemodynamic inference, which has the potential to help overcome the barrier of limited data acquisition in healthcare. 

Furthermore, we focus both on identifying abnormal mPCWP as a binary classification task and predicting the exact mPCWP values as a regression task, using either binary or continuous label-based mining as well as the DML training scheme. This is important, as medical datasets frequently comprise a mix of categorical and continuous data, necessitating both classification and regression tasks for model-based clinical status inference. For example, intracardiac pressure obtained through right heart catheterization is one typical technique in which clinicians seek to see if a patient has abnormal intracardiac pressure (e.g., increased mPCWP) or if the value changes before or after treatment. 
\section{Related Work}
\subsection{Clinical Prediction with ECGs}
 
There is a large body of work developing a deep learning-based ECG classifier for inferring cardiac abnormalities \citep{hannun2019cardiologist, acharya2017application, masetic2016congestive, acharya2019deep}. Recent research has extended beyond ECG classification tasks to regression tasks that infer hemodynamics, such as cardiac output or intracardiac pressure including pulmonary artery pressure and mPCWP \citep{schlesinger2022deep, raghu2023ecg}. Additionally, recent research has shown that ECG signals also encode information about patient demographics by demonstrating age and sex prediction with ECGs \citep{attia2019age}. This shows that leveraging the encoded physiological and demographic information in ECGs could potentially improve model robustness and accuracy.

\paragraph{Contrastive Learning with ECGs and Biomedical Signals} Contrastive learning performs pre-training with pretext tasks to generate a global embedding over an unlabeled dataset, by encouraging the invariance to perturbations in the representations \citep{chen2020simple, he2020momentum, grill2020bootstrap}. There have been several attempts to leverage self-supervised representation learning with unlabeled biomedical signal datasets \citep{zhang2022self, cheng2020subject} including ECGs \citep{diamant2022patient, kiyasseh2021clocs, gopal20213kg, lan2022intra, mehari2022self, oh2022lead, raghu2023sequential} and multimodal time-series data \citep{raghu2023sequential}. However, to the best of our knowledge, self-supervised DML has not been applied to the downstream tasks involving ECGs. In this work, we propose a novel method to use DML on ECG datasets without labels (Self-supervised DML) along with conventional DML methods with labels (Supervised DML) for downstream hemodynamics classification and regression tasks and compare the test performance with supervised learning and contrastive learning baselines.

\subsection{Deep Metric Learning}
Deep Metric Learning (DML) is a similarity-driven learning method for building a representation by learning a distance metric. DML has been demonstrated to improve generalization in representation learning \citep{roth2021simultaneous, milbich2021characterizing, dullerud2021fairness}, including some domain-specific applications to ECG datasets \citep{yu2021automatic, zhu2022dual}. DML works by optimizing the loss function defined on pairs or tuples generated from training examples, where the objective is to preserve the relative distances between pairs of data points in the training data. Triplet loss \citep{wu2017sampling, yu2018correcting} is the representative DML objective function that pulls similar samples together and pushes the dissimilar samples away. Extending this objective, angular loss \citep{wang2017deep} introduces geometric restrictions to achieve a higher convergence rate with scale invariance. Margin loss \citep{wu2017sampling} utilizes the learnable boundary across samples with the same and different labels, which extends the triplet loss.

While contrastive learning defines positive samples to be the perturbed views and negative samples to be the views from other samples, traditional DML defines positive and negative samples based on labels (positive samples are the ones with the same label as an anchor). Negative samples can be mined randomly (random mining \citep{hu2014discriminative}) from the pool of samples with different labels to the anchor, and can also be mined using several hard negative mining techniques. One such technique is semihard mining \citep{schroff2015facenet}, where the hard negatives are selected from the samples that exist within a certain distance margin from the positive sample. Another technique is softhard mining \citep{roth2020revisiting}, where the hard negatives are selected between the maximum margin from positive samples and the minimum distance between the negative and anchor.

As DML generally requires the class of labels to form the tuplets of data, only a few studies have applied DML to regression problems \citep{huang2018regression}. Furthermore, only a few recent works proposed ways for establishing DML without using labels to sample triplets \citep{dutta2020unsupervised, iscen2018mining, dutta2020a_unsupervised}. Here we apply self-supervised DML to a real-world regression problem to determine if similarity information within ECG dataset as well as labels helps create an embedding space that improves the performance of downstream hemodynamics inference. Although there are several self-supervised DML works \citep{paixao2020fast, fu2021self, fu2021deep}, to the best of our knowledge, our study is the first to use distance-based ranking to sample the triplets among the pool of unlabeled pertaining dataset.
\section{Deep Metric Learning for Hemodynamics Inference}
\label{methods}
In this work, we evaluate whether DML methods can be used to build a robust representation of cardiac waveforms by leveraging labeled and unlabeled ECG data. We develop and assess two training strategies: supervised DML for the joint learning of hemodynamics classification and regression, and self-supervised DML pre-training for the downstream hemodynamics inference (classification and regression).

\subsection{Supervised Deep Metric Learning for Hemodynamics Classification and Regression}
\label{supdml_classification}
\paragraph{Problem Setting:} Given a sample size $N$ labeled ECG dataset with N = 8,172 data points: $\mathcal{D}_{sup} = \{(x_i, y_i)| (x_1, y_1), \cdots, (x_N, y_N)\}$. Here, $x_i \in \mathbf{R}^{d \times \mathcal{T}}$ and $y_i \in \{0, 1\}$ with $d$ being the number of ECG leads and $\mathcal{T}$ is the number of ECG timesteps. The binary label $y$ describes whether a patient has elevated mPCWP or not, which is defined by the cutoff mPCWP threshold $18$ mmHg (1 if the patient has elevated mPCWP ($>18$ mmHg)).

\paragraph{Binary Label Based Mining:} For each anchor ECG $x_a$, we define the positive sample $x_p$ to be the ECG having the same label as the anchor ECG and the negative sample $x_n$ is defined to be the one that has a different label. The triplet $(x_a, x_p, x_n)$ are then sampled from the batch $\mathcal{B}$ to calculate the triplet loss $\mathcal{L}_{triplet}$. Negative samples can be mined randomly from the batch of samples with different labels from anchors (\textit{Random Mining}). We further applied hard negative mining methods (\textit{Semihard Mining} \citep{schroff2015facenet} and \textit{Sofhard Mining} \citep{roth2020revisiting}) which select harder negatives around positive samples to build a robust DML representation. We introduce the formulations of hard negative mining in Appendix \ref{hardnegative}. 

\paragraph{Continuous Label Based Mining} (\textit{Label Based Mining})\textbf{:} Given an anchor ECG $x_a$, we mine the triplet $(x_a, x_p, x_n)$ from a batch based on the continuous label mPCWP value, where we designate the positive sample $x_p$ as the ECG with an mPCWP value that is closest to the anchor label. Conversely, we define the negative sample, $x_n$, as the ECG with an mPCWP value that is furthest from the anchor label. The detailed method is in Appendix \ref{labelbased}.

\begin{figure*}[t!]
    \centering
    \includegraphics[width=\textwidth]{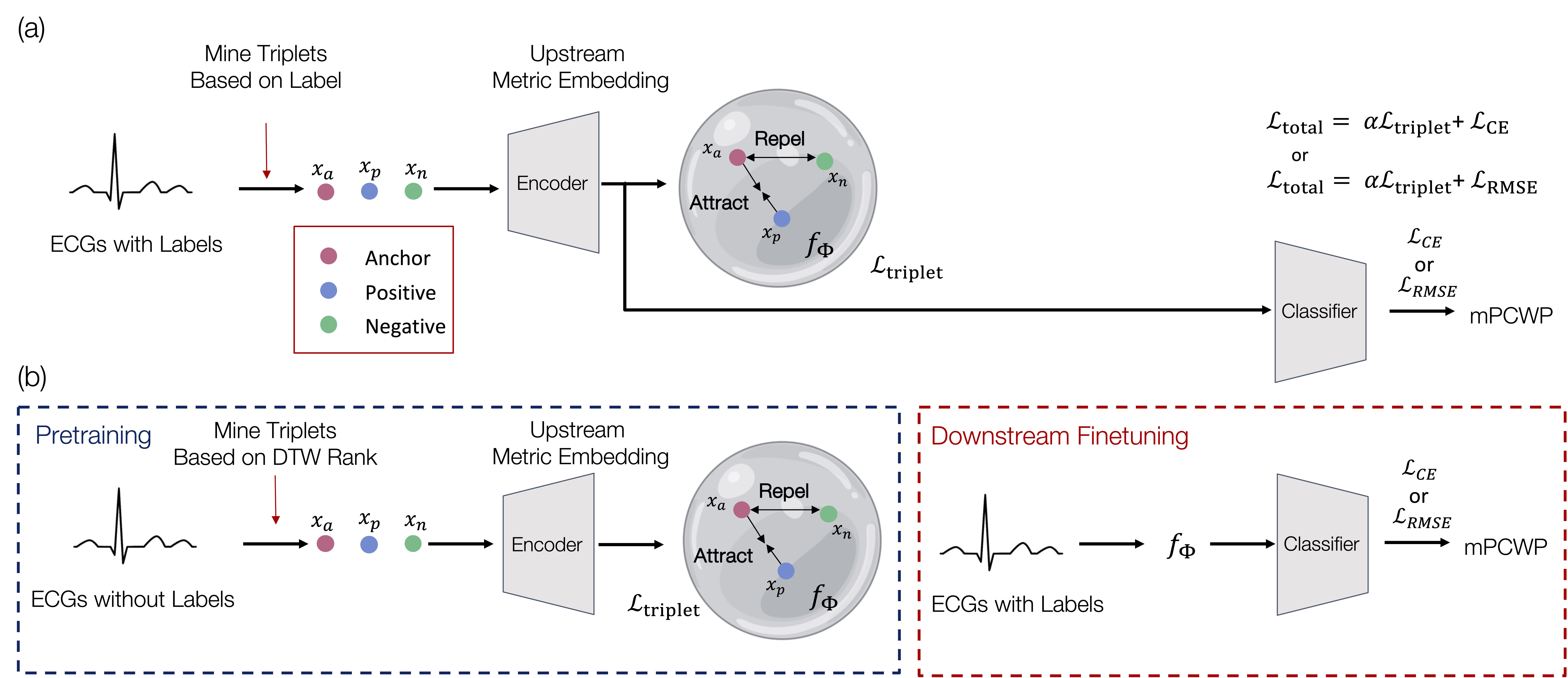}
    \caption{Intracardiac pressure inference using ECGs with DML. \textbf{(a) Supervised DML for Joint-Learning of Hemodynamics Classification and Regression.} ECG triplets are mined based on mPCWP labels (binary label indicating elevation of mPCWP $>$ 18 mmHg), which are used to build the upstream metric embedding $f_\Phi$. Triplet loss from ECG triplets and classification/regression loss are jointly learned in an end-to-end fashion. \textbf{(b) Self-Supervised DML.} Using the unlabeled ECG corpus (ECGs without matched mPCWP), we learn upstream metric embedding $f_\Phi$ with deep metric learning objective (\textit{Pretraining}). Then we finetune this embedding $f_\Phi$ with ECGs with labels to infer mPCWP (\textit{Downstream Finetuning}).}
    \label{concept}
\end{figure*}

\paragraph{Supervised DML Classification:} The model $f_\Phi$ is an encoder has $D$-dimensional output, optimized with triplet loss ($\mathcal{L}_{Triplet} = \sum_{(x_a, x_p, x_n)\in \mathcal{B}} \{||f_\Phi(x_a)-f_\Phi(x_p)|| - ||f_\Phi(x_a)-f_\Phi(x_n)||$). The last fully connected layer $g$ gets the input from the model embedding $f_\Phi$ that has not optimized with triplet loss, and $g \circ f_\Phi$ is optimized by the cross-entropy loss ($\mathcal{L}_{CE}$). The final joint training objective combines the two loss functions, for a batch size of $|\mathcal{B}|$ with the triplet loss scaling factor of $\alpha$: $\mathcal{L}_{tot} = \mathcal{L}_{CE} + \alpha\mathcal{L}_{triplet}$ (Figure \ref{concept} (a), See Appendix \ref{supdml_class} for full loss function). We further experiment with two different objective functions (angular loss and margin loss) in replacement of triplet loss, where we introduce the learnable boundary between positive and negative pairs in margin loss and geometric constraints for angular loss. See Appendix \ref{dml_objectives} for detailed explanations of different loss functions.

\paragraph{Supervised DML Regression:} Using the mined triplets $(x_a, x_p, x_n)$, the model $f_\Phi$ and the final classifier output $g \circ f_\Phi$ are jointly trained in an end-to-end fashion with the objective adding triplet loss and Root Mean Square Error (RMSE) with the triplet scaling factor of $\alpha$: $\mathcal{L}_{tot} = \mathcal{L}_{RMSE} + \alpha\mathcal{L}_{triplet}$. Unlike the classification task above, the regressor output predicts the mPCWP value itself. See Appendix Section \ref{supdml_reg} for the full loss function. For both classification and regression joint training, we performed an exhaustive search to determine the optimal value of the triplet loss scaling factor $\alpha$ to enhance the performance of our model (Table \ref{alpha} in Appendix Section \ref{optimal_alpha}).

\subsection{Self-Supervised Deep Metric Learning}
\label{selfsupdml}
\paragraph{Problem Setting:} Now we extend our approach to self-supervised DML where we have larger body of $M$ = 5.4 million ECG data points without the matched mPCWP labels $y$: $\mathcal{D}_{self-sup} = \{x_j| x_1, \cdots, x_M\}$, where $x_j \in \mathbf{R}^{d \times \mathcal{T}}$.

\paragraph{Distance Based Mining:} Given an anchor $x_a$ in a batch, a positive sample corresponding to each anchor is determined as the one with the smallest distance metric which compares the ECG signal inputs. We sample triplets using a distance-based ranking, where the measured distance between two inputs represents the similarity between inputs. From the calculated measures of dynamic time warping (DTW \citep{muller2007dynamic}, $d_{DTW}$) and Euclidean distance between the pairs of ECG samples in a batch, we sample the triplets based on the distance-based ranking. For instance, given an anchor $x_a$, positive sample $x_p$ is the one with the minimum distance from $x_a$ within the batch: $\mathcal{B}_{pos} = \{x_p|\underset{x_p \in \mathcal{B}}{\arg \min } \text{ } d_{DTW}(x_{p}, x_{a})\}$ for the case where the distance measure is DTW. Then we select the negative data points $x_n$ randomly from the set that excludes the positive sample: $\mathcal{B}_{neg} = \{x_n | x_n \in \mathcal{B}, x_n \neq x_p\}$. Note that DTW has calculated for every pair of input batches as it is not a symmetric true distance and thus does not obey the triangle inequality.

\paragraph{Self-Supervised DML Pretraining and Downstream Hemodynamics Inference:} The model $f_\Phi$ is optimized with the upstream DML with $\mathcal{L}_{triplet}$. (Figure \ref{concept} (a) Pretraining) The input triplets $(x_a, x_p, x_n)$ are mapped on D-dimensional representation space $f_\Phi$, where we finetune the pre-trained self-supervised DML model with the downstream hemodynamic inference task (See Appendix \ref{finetuning}, Figure \ref{concept} (b) Downstream Finetuning). The binary classification task is to infer the elevation of mPCWP with the threshold of $18$ mmHg, and the regression task would be to predict the value of mPCWP.

\section{Experimental Design}
\subsection{Experiments}
To evaluate whether DML methods provide robust prediction on hemodynamics inference tasks (classification, regression), we first compare the performance of the DML models (supervised DML, self-supervised DML) against random initialization, supervised learning, and contrastive learning baselines (Table \ref{result}). To evaluate the quality of the representation, we evaluated the learned representation's ability to capture the samples with the same label within the 1-nearest neighbors using Recall@1 (Table \ref{result}). For supervised DML joint-trained models, we explored the mining methods (random, label-based, semihard, and softhard) and DML objectives (triplet, margin, and angular loss) (Table \ref{mining} and Table \ref{loss}). Finally, we analyzed the performance gaps across different age and gender subgroups to measure subgroup bias in model performance (Table \ref{subgroup}, \ref{agegap}, and Figure \ref{perf_gap}).\footnote{The code used for the experiment is available at \url{https://github.com/mandiehyewon/ssldml}.}

\subsection{Models and Implementation Details} 
\subsubsection{DML Models} 
\paragraph{Supervised DML} We used ResNet18 \citep{he2016deep} backbone as an encoder $f_\Phi$ for generating $D$-dimensional space (with D of 128, 256, 512, 1024) to optimize the objective function, and $f_\Phi$ is fed into the classifier $g$ with two fully-connected (FC) layers with batch normalization, ReLU activation, and a dropout rate of 0.3. We add a sigmoid activation function to the last fully connected layer for the classification task.

\paragraph{Self-Supervised DML} We experimented with two similarity or distance measures (DTW, Euclidean distance) for the similarity-based ranking of triplet mining in self-supervised DML. We used the same ResNet18 \citep{he2016deep} backbone as an encoder $f_\Phi$ to generate the embedding dimension $D$ of 128, 256, 512, and 1024. Classifier ($g$) with the same architecture as SupDML (2 FC layers)  is added after encoder $f_\Phi$ for finetuning.

\subsubsection{Baselines}
\label{baselines}
The baselines we used to compare against supervised DML or self-supervised DML are as below. 

\paragraph{1. Random Initialization} We randomly initialized $f_\Phi$ and fine-tuned the model on the downstream task. This serves as a standard baseline for self-supervised methods.

\paragraph{2. Supervised Models} We train \textit{Supervised classification} and \textit{Supervised Regression} models with ResNet18 \citep{he2016deep} architecture that are trained for hemodynamics binary classification (elevated mPCWP) or regression. 

\paragraph{3. Contrastive Learning Models} We evaluate self-supervised contrastive learning models that are pre-trained with unlabeled ECGs (\textit{ECGs without Labels}) used for finetuning on downstream tasks. Finetuning of the contrastive learning models followed the same step as the downstream hemodynamics inference steps in section \ref{selfsupdml}.

\noindent (1) \textit{SimCLR} \citep{chen2020simple} is a self-supervised learning approach that learns useful representations of images by maximizing the agreement between augmented views of the same image. We used the same encoder $f_\Phi$ as DML models (ResNet18), and the same classifier $g$ except the projection head for calculating the contrastive loss. To reproduce the experiments for SimCLR in \cite{kiyasseh2021clocs}, we used three perturbations: Gaussian noise (tested over noise variance hyperparameter 0.01, 1, 10, 100), flipping along the temporal axis, and flipping along the amplitude axis of ECGs.

\noindent (2) \textit{CLOCS} \citep{kiyasseh2021clocs} is a family of contrastive learning that exploits spatial and temporal invariance of ECG signals. We used Contrastive Multi-segment Coding (CMSC) for our CLOCS baseline, which was the best-performing model in the original paper. 

\noindent (3) \textit{PCLR} \citep{diamant2022patient} is patient-contrastive learning where the ECGs from the same patients were paired for self-supervised pre-training.  We used a pre-trained PCLR model provided by the authors for finetuning, which is 320 dimensions.

\subsubsection{Evaluation}
For evaluating the performance of the mPCWP classification task, we utilized the Area Under the Receiver Operating Characteristic Curve (AUC) and the Area Under the Precision-Recall Curve (APR). Additionally, we adopted Root Mean Squared Error (RMSE) as the performance metric for the mPCWP regression task. To ensure the reliability of our results, we reported the mean and standard deviation of 1,000 bootstraps for both tasks. For statistical analysis of performance and the performance gap between models, we utilized the Kruskal-Wallis tests.

To evaluate the quality of the embedding space, we employed the Recall@k \citep{jegou2010product} performance metric, which quantifies the number of instances with identical labels within the k-nearest neighbor of the sample. Specifically, we set k=1 to assess the proportion of instances with matching labels within the 1-nearest neighbor and report Recall@1 (Table \ref{result}). 

To investigate the difference in task inference quality between demographic groups, we tested pre-trained or trained models listed in Section \ref{baselines} on age and gender subgroups of the test dataset. For the performance gap across gender subgroups, we presented the relative difference in performance (AUC, APR, RMSE) calculated by subtracting the performance of the female group from the performance of the male group (Table \ref{subgroup}). For analyzing the performance gap across age subgroups, we calculated the absolute value of the averaged pairwise relative difference in performance between four distinct age subgroups: 18 $\leq$ age $<$ 35, 35 $\leq$ age $<$ 50, 50 $\leq$ age $<$ 75, 75 $\leq$ age.

The proportion of k-nearest neighbors of the embedding space belonging to the same minority gender group (female) is then calculated to measure the degree of minority subgroup segregation in the representation space, with $k=2, 3, 5$ (Table \ref{knn}). We performed the k-nearest neighbor computation with scikit-learn \citep{scikit-learn} and used seaborn \citep{Waskom2021} for the bar plot.

\subsection{Datasets}
We trained and evaluated the models with two sets of datasets with distinct patient cohorts: \textit{ECGs with Labels} and \textit{ECGs without Labels}. Both datasets include primarily 10 total seconds of trace per patient, where these ECGs were acquired under optimal conditions while the patients were at rest. We utilized the dataset \textit{ECGs with Labels} for supervised DML joint-training (Figure Figure \ref{concept} (a)), supervised models, and for the finetuning of pre-trained models (self-supervised DML, contrastive learning, Figure \ref{concept} (b) Downstream Finetuning). We then used \textit{ECGs without Labels} for self-supervised DML and self-supervised contrastive learning pre-training (Figure \ref{concept} (b) Pretraining). Both \textit{labeled} and \textit{unlabeled} ECG datasets below were obtained as part of an IRB-approved protocol.

\paragraph{Dataset with mPCWP Labels (\textit{ECGs with Labels})} We construct the dataset containing 12 lead ECGs with corresponding right heart catheterization procedures from the data warehouse of Massachusetts General Hospital. Our dataset consists of 8,172 ECGs and right heart catheterization procedure data from 4,051 patients who underwent ECG recordings on the same date they had right heart catheterization from 2010 to 2020. The ECG data contains $10$ seconds of signal data, sampled at a rate of 250 Hz, and the label mPCWP was obtained from the right heart catheterization procedure. We partitioned the dataset into the train, valid, and testing sets, which include 4,680, 1,667, and 1,825 data points, as well as 2,430, 810, and 811 patients, respectively. Our training, validation, and testing sets were partitioned ensuring that no patient overlapped across these different sets. Among 811 test set patients, 344 were female and 467 were male, showing a higher representation of male patients compared to females. We have 20 patients in the age group 18-35, 77 patients in the 35-50 group, 428 in 50-75, and 297 in patients aged over 75. Because age and gender is the primary demographic group available in our dataset, we assessed the fairness of models in different age and gender groupings.

\paragraph{Dataset without mPCWP Labels (\textit{ECGs without Labels})} The larger 12 lead ECG dataset without mPCWP mapping has been provided by the agreement from Massachusetts General Hospital and Brigham and Women's Hospital where we have a total of $5,426,614$ ECGs with $1,195,268$ patients who had ECG recordings while their daytime visit or admission. This dataset includes a more general population of patients and was not only limited to heart failure patients. Of all the ECGs of patients, 917,395 patients have ECG signals exceeding 10 seconds in length. In these particular instances, we selected and isolated 10-second segments from the total ECG signal to generate multiple distinct ECG traces. We implemented this method to ensure a standardized length across all ECG samples in our dataset. We removed patient cohorts overlapping with the \textit{ECG datasets with mPCWP labels} and excluded abnormal ECG signals with continued zero amplitude (mV) at any point of recording at any lead. The other preprocessing and preparation of the ECG dataset follow the same procedure used for the \textit{ECGs with Labels}. 
\begin{table*}[h!]
\footnotesize
 \centering
  \caption{Performance of downstream mPCWP classification task with supervised baseline, supervised DML, and unsupervised DML pretraining according to each embedding dimension (Dim). The supervised baseline model utilized an encoder with 128 dimensions. We report the mean with a standard deviation based on 1,000 bootstraps, and models with the best performance are boldfaced.}
 {
 	\begin{tabular}{c|c|c|cccc}
 		\toprule
  		& & & \multicolumn{2}{c}{Classification} &\multicolumn{1}{c}{Regression} &\multicolumn{1}{c}{Embedding}\\
 		\cmidrule(r){2-2}\cmidrule(r){3-3}\cmidrule(r){4-5}\cmidrule(r){6-6}\cmidrule(r){7-7}
 		& Models & Dim & AUC & APR & RMSE  & Recall@1 \\
 		\midrule
            \multicolumn{2}{c|}{Random Initialization}
            & 128
            & 74.3 $\pm$ 1.9
 		& 55.7 $\pm$ 3.3
 		& 7.88 $\pm$ 0.26
            & 68.4
            \\
            \midrule
            \multirow{5}{*}{Supervised}
 		& Supervised
 		& 128
 		& 78.4 $\pm$ 1.1
 		& 59.5 $\pm$ 2.2
 		& 7.45 $\pm$ 0.15
            & -
 		\\
            \cmidrule{2-7}
		& \multirow{4}{*}{\begin{tabular}[c]{@{}c@{}}Supervised DML\\ (Random Mining)\end{tabular}}
		& 128 
            & 76.7 $\pm$ 1.6
 		& 56.1 $\pm$ 3.3
 		& \textbf{7.15} $\pm$ 0.21 
            & \textbf{69.6}
		\\
 		& & 256
            & 76.5 $\pm$ 1.7
 		& 57.2 $\pm$ 3.2
 		& 7.19 $\pm$ 0.20
            & 66.5
 		\\
  		& & 512
            & 77.1 $\pm$ 1.8
 		& 59.6 $\pm$ 3.2
 		& 7.21 $\pm$ 0.18
            & 67.5
  		\\
            & & 1024 
            & 77.5 $\pm$ 1.7
 		& 58.1 $\pm$ 2.9
 		& 7.20 $\pm$ 0.20
            & 66.3
 		\\
 		\midrule
            \midrule
            \multirow{12}{*}{Self-Supervised}
            & SimCLR
            & 128
 		& 75.0 $\pm$ 1.9
 		& 58.2 $\pm$ 3.2
 		& 7.58 $\pm$ 0.22 
            & 68.6
            \\
            \cmidrule{2-7}
            & CLOCS 
            & 128
            & 74.9 $\pm$ 1.7
 		& 54.0 $\pm$ 3.2
 		& 7.61 $\pm$ 0.22 
            & 68.3
            \\
            \cmidrule{2-7}
            & PCLR
            & 320
            & 75.3 $\pm$ 1.5 
 		& 56.0 $\pm$ 1.5
 		& 10.04 $\pm$ 0.27
            & 63.2
            \\
            \cmidrule{2-7}
		\multicolumn{1}{c|}{} 
            & \multirow{4}{*}{\begin{tabular}[c]{@{}c@{}}Self-supervised DML\\ (DTW)\end{tabular}} 
		& 128 
 		& 77.2 $\pm$ 1.7
 		& \textbf{60.5} $\pm$ 3.1 
 		& 7.65 $\pm$ 0.21 
            & \textbf{69.3}
 		\\
 		& 
 		& 256
 		& 76.4 $\pm$ 1.8 
 		& 57.6 $\pm$ 3.1
 		& 7.69 $\pm$ 0.17
            & \textbf{69.4}
 		\\
  		&
            & 512
            & 75.3 $\pm$ 1.8 
 		& 55.2 $\pm$ 3.3
 		& 7.71 $\pm$ 0.26
            & 67.2
  		\\
            &
            & 1024
 		& 74.8 $\pm$ 1.7 
 		& 55.4 $\pm$ 3.1
 		& 7.70 $\pm$ 0.21
            & 63.8
 		\\
 		\cmidrule{2-7}
		& \multirow{4}{*}{\begin{tabular}[c]{@{}c@{}}Self-supervised DML\\ (Euclidean)\end{tabular}}
		& 128
 		& 77.4 $\pm$ 1.8 
 		& 58.2 $\pm$ 3.5
 		& 7.80 $\pm$ 0.22
            & 64.2
 		\\
 		&
 		& 256
 		& 77.2 $\pm$ 1.7 
 		& \textbf{60.7} $\pm$ 3.2
 		& 7.59 $\pm$ 0.20 
            & 61.4
 		\\
  		&
  		& 512
 		&\textbf{79.4 $\pm$ 1.1} 
 		& 59.9 $\pm$ 2.2
 		& 7.64 $\pm$ 0.17 
            & 67.2
  		\\
            &
            & 1024
 		& 76.3 $\pm$ 1.6 
 		& 58.5 $\pm$ 3.3
 		& 7.70 $\pm$ 0.24
            & 67.7
 		\\
 		\bottomrule
 	\end{tabular}
 	}
 \label{result}
\end{table*}

\section{Results}
\subsection{Supervised and Self-supervised DML on Hemodynamics Inference}
We summarize the test performance of DML models and baselines on and downstream hemodynamics inference task in Table \ref{result}. We find that training with DML showed significantly improved performance compared to the baselines. Supervised DML (best RMSE of 7.15 with embedding dimension 128) significantly outperformed random initialization (7.88), supervised (7.38), and contrastive baselines (lowest 7.58 with SimCLR) on the regression task. The regression performance of supervised DML on every embedding dimension (128, 256, 512, 1024) achieved significantly low ($p < 0.0001$) RMSE loss compared to any other baselines. Supervised DML exhibited comparable performance on classification using the APR metric (59.6 with embedding dimension 1024), where its performance was not significantly different from that of the supervised baseline (59.5), while the AUC was significantly higher for the supervised baseline (78.4 for supervised and 77.5 for supervised DML with embedding dimension 1024). This implies that the underlying similarity information in the ECG signal pattern captured by the labels helps infer the downstream wedge pressure value. Supervised DML results were reported with the best triplet loss scaling factor $\alpha$ after hyperparameter tuning on various $\alpha$ values (0.1, 1.0, 2.0, 3.0, 10.0): $\alpha$ of 3.0 and 2.0 were used for classification and regression, respectively, for the embedding dimension of 128. For detailed results on other embedding dimensions (256, 512, 1024) please refer to Table \ref{alpha} in Appendix Section \ref{appendix_supdml}.

We evaluated the classification performance of self-supervised DML and compared it to that of contrastive learning baselines. Our findings show that both DTW-based and Euclidean-based ranking in self-supervised DML achieved significantly better classification performance (Best AUC of 79.4 in supervised DML with Euclidean distance, APR of 60.5 for self-supervised DML with DTW, and 60.7 for Euclidean) compared to the contrastive learning baselines (best performance AUC 75.3 with PCLR and APR of 58.2 with SimCLR) and random initialization (AUC 74.3 and APR 55.7) ($p < 0.0001$ against all baselines). On the other hand, we found no significant difference in regression performance between the contrastive learning baselines (lowest RMSE loss with SimCLR, 7.58) and self-supervised methods (lowest with Euclidean-based ranking, 7.59), while observing a significant improvement over PCLR (10.04) and random initialization (7.88) ($p < 0.0001$). Overall, self-supervised DML demonstrated the optimal trade-off between classification and regression tasks, achieving the best performance in the classification task while maintaining a comparable level of performance in the regression task.

\subsection{Evaluation of the Representation Quality} 
To evaluate the discriminative ability of representation space, we compare the label consistency of close samples using the Recall@1 metric (Table \ref{result}). The best performance was observed in supervised DML with random mining and softhard mining (Table \ref{mining}), as well as self-supervised DML with DTW-based ranking. On the other hand, supervised DML with label based mining, semihard mining (Table \ref{mining}), and self-supervised DML with the Euclidean-based ranking (Table \ref{result}) showed worse performance compared to the contrastive learning (SimLCR and CLOCS). Furthermore, increasing the embedding dimension improved joint-learned task performance in supervised DML (Table \ref{result}), which is consistent with prior DML studies \citep{roth2020revisiting}. However, the increase in embedding dimension did not affect the downstream task performance in supervised DML with hard negative mining (Table \ref{mining} in Appendix Section \ref{mining_appendix}) and self-supervised DML methods (Table \ref{result}). 

\subsection{DML Models Reduce Subgroup Performance Gaps}
\begin{figure*}[h!]
    \centering
    \includegraphics[width=\textwidth]{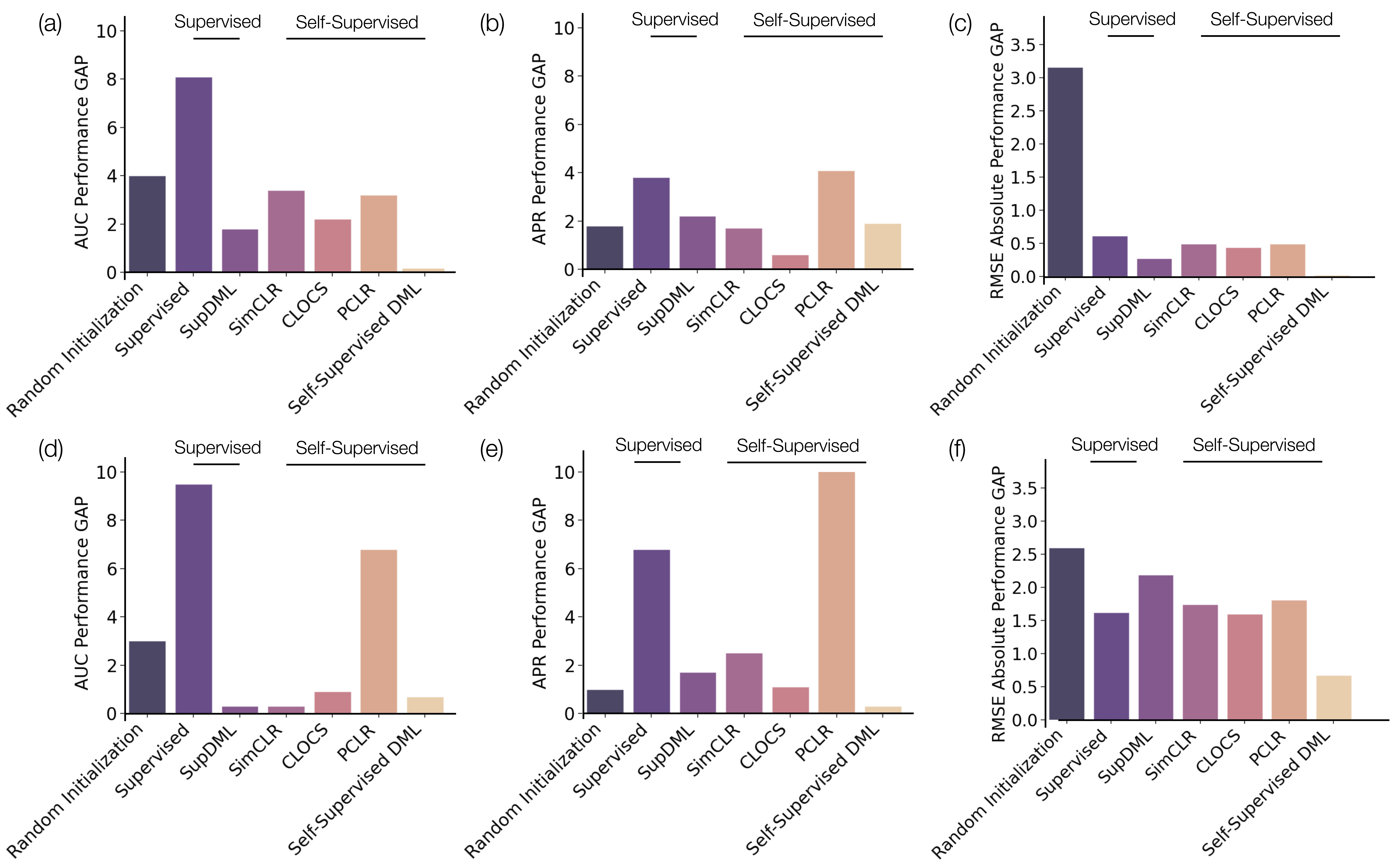}
    \caption{The absolute performance gap of mPCWP inference tasks (classification ((a), (d) AUC, (b), (e) APR) and regression ((c), (f) RMSE)) across demographic subgroups ((a)-(c) gender subgroup and (d)-(f) age subgroup). 
    }
    \label{perf_gap}
\end{figure*}
\begin{table*}[t!]
\footnotesize
 \centering
  \caption{Gender subgroup performance gap of downstream mPCWP classification task has decreased on either Supervised or Self-Supervised DML compared to supervised and contrastive baselines (an embedding dimension of 128 used for DML and contrastive learning encoder). We report the mean with a standard deviation based on 1,000 bootstraps, and models with the lowest performance gaps are boldfaced.}
 {
 	\begin{tabular}{cccccc}
 		\toprule
  		& & & \multicolumn{2}{c}{Classification} &\multicolumn{1}{c}{Regression}\\
 		\cmidrule(r){4-5}\cmidrule(r){6-6}
 		\multicolumn{2}{c}{Models} & Subgroup & AUC & APR & RMSE  \\
 		\midrule
 		\midrule
 		\multicolumn{2}{c}{\multirow{4}{*}{Random Initialization}}
            & Full
            & 74.3 $\pm$ 1.9
 		& 55.7 $\pm$ 3.3
 		& 7.88 $\pm$ 0.26
 		\\
            &
            & Male
 		& 77.3 $\pm$ 1.5
 		& 54.3 $\pm$ 3.1
 		& 8.68 $\pm$ 0.80
  		\\
            &		
            & Female
 		& 73.4 $\pm$ 1.4
 		& 56.1 $\pm$ 2.3
 		& 11.84 $\pm$ 2.52
  		\\
  		&
            & Gap
 		& 4.0
 		& 1.8
 		& 3.16
  		\\
            \midrule
 		\multirow{8}{*}{Supervised}
            & \multirow{4}{*}{Supervised Baseline}
 		& Full
 		& 78.4 $\pm$ 1.1
 		& 59.5 $\pm$ 2.2
 		& 7.45 $\pm$ 0.15
 		\\
            & 
  		& Male
 		& 81.3 $\pm$ 1.4
 		& 61.6 $\pm$ 2.9
 		& 7.17 $\pm$ 0.18
  		\\
            &
  		& Female
 		& 73.2 $\pm$ 2.0
 		& 57.8 $\pm$ 3.5
 		& 7.88 $\pm$ 0.29
  		\\
            &
  		& Gap
 		& 8.1
 		& 3.8
 		& 0.71
  		\\
            \cmidrule{2-6}
            & \multirow{4}{*}{\begin{tabular}[c]{@{}c@{}}Supervised DML\\ (Random Mining)\end{tabular}}
 		& Full
            & 76.7 $\pm$ 1.6
 		& 56.1 $\pm$ 3.3
 		& 7.15 $\pm$ 0.21
 		\\
            &
 		& Male
  		& 77.4 $\pm$ 1.5
 		& 54.2 $\pm$ 2.8
 		& 7.07 $\pm$ 0.19
 		\\
            &
  		& Female
 		& 75.6 $\pm$ 1.2
 		& 56.4 $\pm$ 2.4
 		& 7.34 $\pm$ 0.17
  		\\
            &
  		& Gap
 		& \textbf{1.8}
 		& \textbf{2.2}
 		& \textbf{0.27}
            \\
            \midrule
            \midrule
            \multirow{16}{*}{Self-Supervised}
 		& \multirow{4}{*}{SimCLR}
 		& Full
 		& 75.0 $\pm$ 1.9
 		& 58.2 $\pm$ 3.2
 		& 7.58 $\pm$ 0.22
 		\\
  		&
  		& Male
 		& 76.2 $\pm$ 1.7
 		& 54.8 $\pm$ 3.2
 		& 7.18 $\pm$ 0.17
  		\\
            &
  		& Female
 		& 72.8 $\pm$ 1.4
 		& 56.5 $\pm$ 2.4
            & 7.67 $\pm$ 0.16
  		\\
            &
  		& Gap
 		& 3.4
 		& 1.7
 		& 0.49
  		\\
            \cmidrule{2-6}            
 		& \multirow{4}{*}{CLOCS}
 		& Full
            & 75.9 $\pm$ 1.7
 		& 56.2 $\pm$ 3.4
 		& 7.45 $\pm$ 0.22
 		\\
  		&
  		& Male
 		& 76.5 $\pm$ 1.6
 		& 55.5 $\pm$ 2.9
 		& 7.22 $\pm$ 0.17
  		\\
            &
  		& Female 
 		& 74.3 $\pm$ 1.4
 		& 56.1 $\pm$ 2.3
 		& 7.66 $\pm$ 0.17
  		\\
            &
  		& Gap
 		& 2.2
 		& \textbf{0.6}
 		& 0.44
  		\\
            \cmidrule{2-6}
 		& \multirow{4}{*}{PCLR}
 		& Full
            & 75.3 $\pm$ 1.5
 		& 56.0 $\pm$ 2.4
 		& 10.04 $\pm$ 0.27
 		\\
  		&
  		& Male 
            & 75.1 $\pm$ 1.5
 		& 53.9 $\pm$ 2.3
 		& 11.14 $\pm$ 0.24
  		\\   
            &
  		& Female
 		& 71.9 $\pm$ 1.7
 		& 58.0 $\pm$ 2.7
 		& 11.63 $\pm$ 0.34
  		\\
            &
  		& Gap
 		& 3.2
 		& 4.1
 		& 0.49
  		\\
            \cmidrule{2-6}          
 		& \multirow{4}{*}{\begin{tabular}[c]{@{}c@{}}Self-supervised DML\\ (DTW)\end{tabular}}
 		& Full
 		& 77.2 $\pm$ 1.7
 		& 60.5 $\pm$ 3.1
 		& 7.65 $\pm$ 0.21
 		\\
  		&
  		& Male
 		& 75.0 $\pm$ 1.0
 		& 57.1 $\pm$ 3.4
 		& 7.82 $\pm$ 0.20
  		\\
            &
            & Female
 		& 75.0 $\pm$ 1.3
 		& 59.0 $\pm$ 2.3
 		& 7.84 $\pm$ 0.17
            \\
           &
  		& Gap
 		& \textbf{0.0} 
 		& 1.9
 		& \textbf{0.02} 
 		\\
 		\bottomrule
 	\end{tabular}
 	}
 \label{subgroup}
\end{table*}

We investigated the differences in mPCWP inference among demographic groups (Table \ref{subgroup}, \ref{agegap}, and Figure \ref{perf_gap}). For the gender groups, we reported the absolute performance difference between male and female groups of 1,000 bootstraps (Table \ref{subgroup}, rows 'Gap'), and for the age groups, we calculated the average pairwise difference in performance between any two groups. These performance gaps are relative differences in performance. 

Supervised and self-supervised DML helped achieve a significantly fairer prediction of downstream tasks across different demographic subgroups against the baselines. Among supervised models (supervised baseline, supervised DML), supervised DML showed low gaps for all the performance metrics (AUC, APR, and RMSE) compared to the supervised baseline and random initialization, achieving fair prediction across different gender subgroups. Supervised DML also achieved lower AUC and RMSE gaps compared to random initialization and self-supervised contrastive learning baselines (SimCLR, CLOCS, PCLR). Self-supervised learning showed the lowest performance gap on both AUC and RMSE with statistical significance with any other baselines ($p < 0.0001$), and SimCLR and CLOCS achieved a lower performance gap on APR compared to supervised and self-supervised DML. A comprehensive discussion of the performance gaps across age subgroups can be found in Appendix Section \ref{appendix_agegap}.

The improvements in fair classification and regression imply that the pre-trained embedding learned with DML objectives, either using the triplets sampled based on label (supervised DML) or distance-based similarity metric (self-supervised DML), represent the physiological difference related to demographic groups, supporting the previous work that ECG signal can provide information to infer demographic groups \citep{attia2019age}. 

\subsection{Impact of varying DML Objectives}
\label{dmlobjectives}

In Sections \ref{dmlobjectives} and \ref{dmlmining}, we explore the effect of varying DML objectives and mining techniques, and the results show that our supervised DML model shows robust performance and achieves equitable prediction. 

\begin{figure*}[!ht]
    \centering
    \includegraphics[width=1.0\textwidth, keepaspectratio]{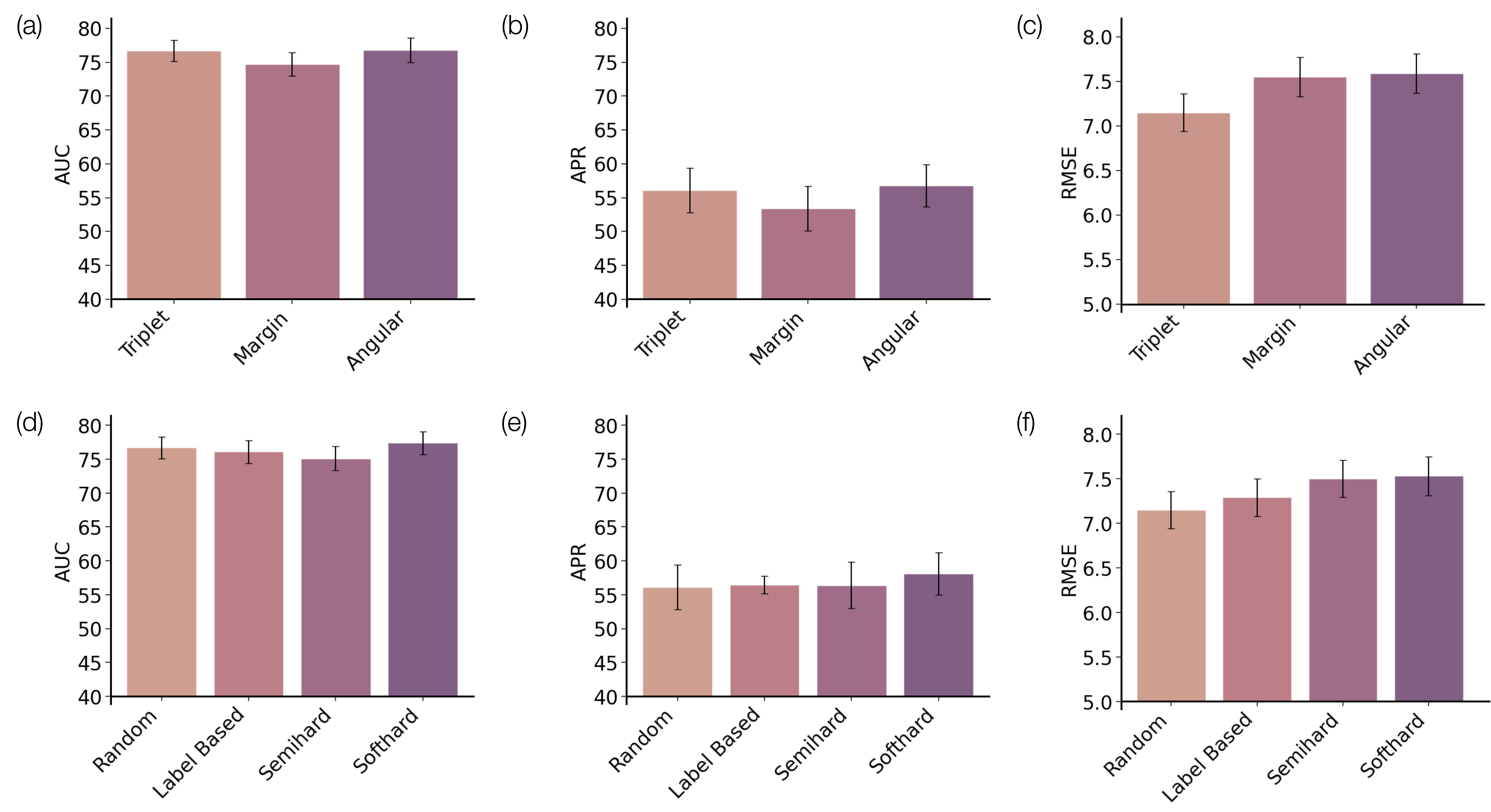}
    \caption{Evaluation of performance on various DML objectives ((a)-(c)) and mining strategies ((d)-(f)). AUC (a), APR (b), and RMSE of triplet, margin, and angular loss (c). AUC (d), APR (e), and RMSE of random, semihard, and softhard mining.}
    \label{ablation}
\end{figure*}

We investigated the impact of different DML objectives on the classification and regression tasks of mPCWP inference (Figure \ref{ablation} (a)-(c)). We utilized margin loss and angular loss, along with the triplet loss, and trained them jointly with the loss for mPCWP inference (as described in Section \ref{supdml_classification}). A detailed description of the objectives can be found in Appendix Section \ref{dml_objectives}. Our results showed that the angular loss performed nonsignificantly similar to the triplet loss on the classification task, significantly outperforming the performance with margin loss (Table \ref{loss}). However, the triplet loss outperformed the angular loss in terms of RMSE on the regression task, for the supervised DML method.

\subsection{Impact of varying Supervised DML Mining}
\label{dmlmining}
We evaluated the performance of label based mining and different hard negative mining methods (semihard and softhard mining) against the random mining method in supervised DML for mPCWP elevation classification and regression tasks across all embedding dimensions. The results are summarized in Figure \ref{ablation} and Table \ref{mining} of Appendix \ref{mining_appendix}. We found that label based mining with 1024 dimensions showed AUC performance on par with random mining. Furthermore, softhard mining with 128 dimensions showed comparable classification performance to the result with random mining, while semihard mining showed regression performance similar to random mining. Interestingly, the trend of classification performance improvement with increased embedding dimension, observed in random mining, was not clearly visible in the explored label based mining and hard negative mining methods (semihard and softhard). This result indicates that the impact of embedding dimension on performance is not consistent when using label based or hard negative mining methods. 

\paragraph{Subgroup Performance with Label Based Mining and Hard Negative Mining} We report the subgroup performance gap of Supervised DML joint trained with label based and hard negative mining methods in Figure \ref{mining_gap} (d)-(f) and Table \ref{mining_subgroup} of Appendix \ref{mining_appendix}. Our results indicate that although softhard mining showed a higher performance gap compared to random mining in the classification task, the performance gap of APR and RMSE has decreased in Semihard mining. Also, label based mining showed the lowest gap in AUC and showed gaps in other performance metrics comparable to random mining. This implies that Supervised DML embedding built with triplets with harder negatives helps achieve fairer prediction, while not harming the performance too much. The detailed performance report is on Table \ref{mining_subgroup} of Appendix \ref{mining_appendix}.

\begin{figure*}[h!]
    \centering
    \includegraphics[width=\textwidth]{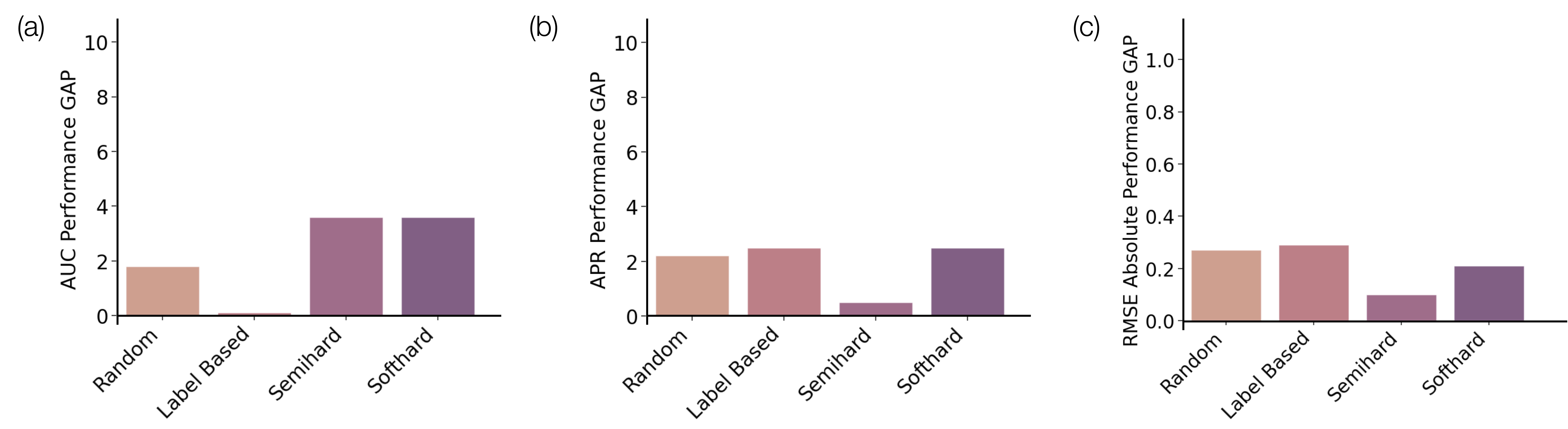}
    \caption{Effect of supDML hard negative mining on performance gap. Performance gap of male and female test subgroups on mPCWP inference tasks (Classification ((a) AUC, (b) APR) and Regression ((c) RMSE)) are reported with a bar plot. A positive value indicates that the male group performed better than the female group, and a negative value indicates the other way around. We report the absolute difference for (c) RMSE.}
    \label{mining_gap}
\end{figure*}
\section{Discussion}
In this work, we proposed a DML approach for the robust representations of electrocardiogram (ECG) signals for the hemodynamics inference task. Our proposed DML approach significantly outperforms random initialization, supervised baseline and existing contrastive learning methods, where supervised DML achieved the best regression performance and self-supervised DML methods achieved the best classification performance. This result demonstrates its ability to learn contextual representations preexisting in ECGs for better generalizability and robustness. More importantly, we showed with our experiment that supervised and self-supervised DML achieves the fair inference of mPCWP across different age and gender subgroups, which significantly improves the subgroup performance gaps in contrastive and supervised baselines. Our proposed method captures the changes in hemodynamics based on individual medical history, promising a potential to be applied for personalized physiological tracking. Our work highlights the potential of DML approaches for improving inference using ECG signals and other biomedical signals.

In our analysis of fairness in supervised and self-supervised DML models, we have observed some interesting trends (Table \ref{knn}). In the supervised baseline and contrastive learning baselines (PCLR, SimCLR, CLOCS), a majority of instances in the minority gender group (Female) are surrounded by instances of the same gender group, indicating less uniformity in the embedding space. However, this trend is less apparent in both supervised and self-supervised DML, where fewer same-gender instances are found in the k-nearest neighbors of the minority group (Female). Supervised and self-supervised DML models have fewer instances of same-gender subgroups surrounding the data points (e.g., female groups are surrounded by 68.21\% of the same gender in 2-NN for supervised DML. Females are surrounded by 69.27\% same-gender instances in self-supervised DML). This could be an explanation for the decreased performance disparity between subgroups in supervised and self-supervised DML models.

\begin{table*}[h!]
\footnotesize
 \centering
  \caption{The proportion (\%) of k-nearest neighbors of the embedding space belonging to the same gender with k = 2, 3, and 5, to assess the degree of gender-based subgroup segregation in the representation space.} 
 {
 	\begin{tabular}{c|ccc}
 		\toprule
 		Models & k=2 & k=3 & k=5\\
            \midrule
		Supervised Baseline
 		& 81.08
 		& 73.84
            & 65.18
            \\
		Supervised DML
 		& 68.21
 		& 59.31
            & 51.81
  		\\
 		\midrule
            PCLR
 		& 69.87
 		& 60.18
            & 51.21
            \\
            SimCLR
 		& 71.73
 		& 60.71
            & 51.84
            \\
            CLOCS
 		& 73.84
 		& 64.37
            & 56.01
            \\
		Self-Supervised DML
 		& 69.27
 		& 58.60
            & 50.80
            \\
 		\bottomrule
 	\end{tabular}
 	}
 \label{knn}
\end{table*}

As DML effectively leverages the notion of similarity, it improves the quality of embedding by locating similar samples together although not explicitly ensuring fair prediction. In the context of fairness in DML, previous studies have made significant strides. \cite{ilvento2019metric} proposed a method for a metric approximation that specifically targets individual fairness. On the other hand, \cite{dullerud2021fairness} pioneered new fairness metrics in DML, and proposed a way of de-correlating the propagated bias in DML representation. Our proposed DML approach also pushed toward fairness in DML. Supervised and self-supervised DML not only provided robust representations for ECG signals in hemodynamics inference tasks but also demonstrated improved fairness in gender subgroup predictions. Our results suggest that DML models are capable of achieving uniformity of demographics in the embedding space, and this could help mitigate subgroup performance gaps, contributing to more equitable inference. These findings support the application of DML approaches for ECG signals and other biomedical signals, ultimately promoting fair and personalized physiological surveillance for diverse patient populations.

\paragraph{Limitations} Despite the promising results, our study has several limitations. First, our analysis relies on internal validation. Leveraging the secondary source of ECG data with matched mPCWP for external validation can further solidify the application of DML models on mPCWP inference. Second, we have only focused on a limited number of problems related to intracardiac pressure classification and regression. Further research is needed to investigate the effectiveness of DML pretraining for other types of downstream tasks detecting cardiac conditions such as arrhythmia \citep{hannun2019cardiologist}, myocardial infarction \citep{acharya2017application}, and heart failure \citep{masetic2016congestive, acharya2019deep}. The generalizability of the DML pre-trained model can be further validated by finetuning it on the public benchmark datasets with larger labeled data corpus, such as PhysioNet 2021 \cite{reyna2021will}, Chapman \citep{zheng2020optimal, zheng202012}, and PTB-XL \citep{wagner2020ptb} datasets. Auditing the models on different subgroups (e.g., self-reported racial groups, insurance plans) available in those datasets would help validate the fairness gain of the model. Furthermore, we can assess the intersections of various sensitive features \citep{morina2019auditing} to gain greater insight into achieving equitable prediction of the model.
\section{Acknowledgements}
This research was supported in part by funds from Quanta Computer, Inc. We would like to thank members of the Computational Cardiovascular Research Group and HealthyML Lab at MIT, and the reviewers for their invaluable feedback.


\bibliography{reference}

\begin{thebibliography}{60}
\providecommand{\natexlab}[1]{#1}
\providecommand{\url}[1]{\texttt{#1}}
\expandafter\ifx\csname urlstyle\endcsname\relax
  \providecommand{\doi}[1]{doi: #1}\else
  \providecommand{\doi}{doi: \begingroup \urlstyle{rm}\Url}\fi

\bibitem[Abraham et~al.(2011)Abraham, Adamson, Bourge, Aaron, Costanzo,
  Stevenson, Strickland, Neelagaru, Raval, Krueger,
  et~al.]{abraham2011wireless}
William~T Abraham, Philip~B Adamson, Robert~C Bourge, Mark~F Aaron, Maria~Rosa
  Costanzo, Lynne~W Stevenson, Warren Strickland, Suresh Neelagaru, Nirav
  Raval, Steven Krueger, et~al.
\newblock Wireless pulmonary artery haemodynamic monitoring in chronic heart
  failure: a randomised controlled trial.
\newblock \emph{The Lancet}, 377\penalty0 (9766):\penalty0 658--666, 2011.

\bibitem[Abraham et~al.(2016)Abraham, Stevenson, Bourge, Lindenfeld, Bauman,
  and Adamson]{abraham2016sustained}
William~T Abraham, Lynne~W Stevenson, Robert~C Bourge, Jo~Ann Lindenfeld,
  Jordan~G Bauman, and Philip~B Adamson.
\newblock Sustained efficacy of pulmonary artery pressure to guide adjustment
  of chronic heart failure therapy: complete follow-up results from the
  champion randomised trial.
\newblock \emph{The Lancet}, 387\penalty0 (10017):\penalty0 453--461, 2016.

\bibitem[Acharya et~al.(2017)Acharya, Fujita, Oh, Hagiwara, Tan, and
  Adam]{acharya2017application}
U~Rajendra Acharya, Hamido Fujita, Shu~Lih Oh, Yuki Hagiwara, Jen~Hong Tan, and
  Muhammad Adam.
\newblock Application of deep convolutional neural network for automated
  detection of myocardial infarction using ecg signals.
\newblock \emph{Information Sciences}, 415:\penalty0 190--198, 2017.

\bibitem[Acharya et~al.(2019)Acharya, Fujita, Oh, Hagiwara, Tan, Adam, and
  Tan]{acharya2019deep}
U~Rajendra Acharya, Hamido Fujita, Shu~Lih Oh, Yuki Hagiwara, Jen~Hong Tan,
  Muhammad Adam, and Ru~San Tan.
\newblock Deep convolutional neural network for the automated diagnosis of
  congestive heart failure using ecg signals.
\newblock \emph{Applied Intelligence}, 49:\penalty0 16--27, 2019.

\bibitem[Attia et~al.(2019)Attia, Friedman, Noseworthy, Lopez-Jimenez, Ladewig,
  Satam, Pellikka, Munger, Asirvatham, Scott, et~al.]{attia2019age}
Zachi~I Attia, Paul~A Friedman, Peter~A Noseworthy, Francisco Lopez-Jimenez,
  Dorothy~J Ladewig, Gaurav Satam, Patricia~A Pellikka, Thomas~M Munger,
  Samuel~J Asirvatham, Christopher~G Scott, et~al.
\newblock Age and sex estimation using artificial intelligence from standard
  12-lead ecgs.
\newblock \emph{Circulation: Arrhythmia and Electrophysiology}, 12\penalty0
  (9):\penalty0 e007284, 2019.

\bibitem[Baudry et~al.(2022)Baudry, Coutance, Dorent, Bauer, Blanchart,
  Boignard, Chabanne, Delmas, d'Ostrevy, Epailly, et~al.]{baudry2022prognosis}
Guillaume Baudry, Guillaume Coutance, Richard Dorent, Fabrice Bauer, Katrien
  Blanchart, Aude Boignard, C{\'e}line Chabanne, Cl{\'e}ment Delmas, Nicolas
  d'Ostrevy, Eric Epailly, et~al.
\newblock Prognosis value of forrester's classification in advanced heart
  failure patients awaiting heart transplantation.
\newblock \emph{ESC Heart Failure}, 9\penalty0 (5):\penalty0 3287--3297, 2022.

\bibitem[Burkhoff et~al.(2003)Burkhoff, Maurer, and Packer]{burkhoff2003heart}
Daniel Burkhoff, Mathew~S Maurer, and Milton Packer.
\newblock Heart failure with a normal ejection fraction: is it really a
  disorder of diastolic function?, 2003.

\bibitem[Chen et~al.(2020)Chen, Kornblith, Norouzi, and Hinton]{chen2020simple}
Ting Chen, Simon Kornblith, Mohammad Norouzi, and Geoffrey Hinton.
\newblock A simple framework for contrastive learning of visual
  representations.
\newblock In \emph{International conference on machine learning}, pages
  1597--1607. PMLR, 2020.

\bibitem[Cheng et~al.(2020)Cheng, Goh, Dogrusoz, Tuzel, and
  Azemi]{cheng2020subject}
Joseph~Y Cheng, Hanlin Goh, Kaan Dogrusoz, Oncel Tuzel, and Erdrin Azemi.
\newblock Subject-aware contrastive learning for biosignals.
\newblock \emph{arXiv preprint arXiv:2007.04871}, 2020.

\bibitem[Diamant et~al.(2022)Diamant, Reinertsen, Song, Aguirre, Stultz, and
  Batra]{diamant2022patient}
Nathaniel Diamant, Erik Reinertsen, Steven Song, Aaron~D Aguirre, Collin~M
  Stultz, and Puneet Batra.
\newblock Patient contrastive learning: A performant, expressive, and practical
  approach to electrocardiogram modeling.
\newblock \emph{PLoS Computational Biology}, 18\penalty0 (2):\penalty0
  e1009862, 2022.

\bibitem[Drazner et~al.(2008)Drazner, Hellkamp, Leier, Shah, Miller, Russell,
  Young, Califf, and Nohria]{drazner2008value}
Mark~H Drazner, Anne~S Hellkamp, Carl~V Leier, Monica~R Shah, Leslie~W Miller,
  Stuart~D Russell, James~B Young, Robert~M Califf, and Anju Nohria.
\newblock Value of clinician assessment of hemodynamics in advanced heart
  failure: the escape trial.
\newblock \emph{Circulation: Heart Failure}, 1\penalty0 (3):\penalty0 170--177,
  2008.

\bibitem[Dullerud et~al.(2021)Dullerud, Roth, Hamidieh, Papernot, and
  Ghassemi]{dullerud2021fairness}
Natalie Dullerud, Karsten Roth, Kimia Hamidieh, Nicolas Papernot, and Marzyeh
  Ghassemi.
\newblock Is fairness only metric deep? evaluating and addressing subgroup gaps
  in deep metric learning.
\newblock In \emph{International Conference on Learning Representations}, 2021.

\bibitem[Dutta et~al.(2020{\natexlab{a}})Dutta, Harandi, and
  Sekhar]{dutta2020a_unsupervised}
Ujjal~Kr Dutta, Mehrtash Harandi, and C~Chandra Sekhar.
\newblock Unsupervised metric learning with synthetic examples.
\newblock In \emph{Proceedings of the AAAI Conference on Artificial
  Intelligence}, volume~34, pages 3834--3841, 2020{\natexlab{a}}.

\bibitem[Dutta et~al.(2020{\natexlab{b}})Dutta, Harandi, and
  Sekhar]{dutta2020unsupervised}
Ujjal~Kr Dutta, Mehrtash Harandi, and Chellu~Chandra Sekhar.
\newblock Unsupervised deep metric learning via orthogonality based
  probabilistic loss.
\newblock \emph{IEEE Transactions on Artificial Intelligence}, 1\penalty0
  (1):\penalty0 74--84, 2020{\natexlab{b}}.

\bibitem[Fu et~al.(2021{\natexlab{a}})Fu, Li, Mao, Wang, and Zhang]{fu2021deep}
Zheren Fu, Yan Li, Zhendong Mao, Quan Wang, and Yongdong Zhang.
\newblock Deep metric learning with self-supervised ranking.
\newblock In \emph{Proceedings of the AAAI Conference on Artificial
  Intelligence}, volume~35, pages 1370--1378, 2021{\natexlab{a}}.

\bibitem[Fu et~al.(2021{\natexlab{b}})Fu, Mao, Yan, Liu, Xie, and
  Zhang]{fu2021self}
Zheren Fu, Zhendong Mao, Chenggang Yan, An-An Liu, Hongtao Xie, and Yongdong
  Zhang.
\newblock Self-supervised synthesis ranking for deep metric learning.
\newblock \emph{IEEE Transactions on Circuits and Systems for Video
  Technology}, 32\penalty0 (7):\penalty0 4736--4750, 2021{\natexlab{b}}.

\bibitem[Gopal et~al.(2021)Gopal, Han, Raghupathi, Ng, Tison, and
  Rajpurkar]{gopal20213kg}
Bryan Gopal, Ryan~W Han, Gautham Raghupathi, Andrew~Y Ng, Geoffrey~H Tison, and
  Pranav Rajpurkar.
\newblock 3kg: Contrastive learning of 12-lead electrocardiograms using
  physiologically-inspired augmentations.
\newblock \emph{arXiv preprint arXiv:2106.04452}, 2021.

\bibitem[Grill et~al.(2020)Grill, Strub, Altch{\'e}, Tallec, Richemond,
  Buchatskaya, Doersch, Avila~Pires, Guo, Gheshlaghi~Azar,
  et~al.]{grill2020bootstrap}
Jean-Bastien Grill, Florian Strub, Florent Altch{\'e}, Corentin Tallec, Pierre
  Richemond, Elena Buchatskaya, Carl Doersch, Bernardo Avila~Pires, Zhaohan
  Guo, Mohammad Gheshlaghi~Azar, et~al.
\newblock Bootstrap your own latent-a new approach to self-supervised learning.
\newblock \emph{Advances in neural information processing systems},
  33:\penalty0 21271--21284, 2020.

\bibitem[Hannun et~al.(2019)Hannun, Rajpurkar, Haghpanahi, Tison, Bourn,
  Turakhia, and Ng]{hannun2019cardiologist}
Awni~Y Hannun, Pranav Rajpurkar, Masoumeh Haghpanahi, Geoffrey~H Tison, Codie
  Bourn, Mintu~P Turakhia, and Andrew~Y Ng.
\newblock Cardiologist-level arrhythmia detection and classification in
  ambulatory electrocardiograms using a deep neural network.
\newblock \emph{Nature medicine}, 25\penalty0 (1):\penalty0 65--69, 2019.

\bibitem[He et~al.(2016)He, Zhang, Ren, and Sun]{he2016deep}
Kaiming He, Xiangyu Zhang, Shaoqing Ren, and Jian Sun.
\newblock Deep residual learning for image recognition.
\newblock In \emph{Proceedings of the IEEE conference on computer vision and
  pattern recognition}, pages 770--778, 2016.

\bibitem[He et~al.(2020)He, Fan, Wu, Xie, and Girshick]{he2020momentum}
Kaiming He, Haoqi Fan, Yuxin Wu, Saining Xie, and Ross Girshick.
\newblock Momentum contrast for unsupervised visual representation learning.
\newblock In \emph{Proceedings of the IEEE/CVF conference on computer vision
  and pattern recognition}, pages 9729--9738, 2020.

\bibitem[Heidenreich et~al.(2022)Heidenreich, Bozkurt, Aguilar, Allen, Byun,
  Colvin, Deswal, Drazner, Dunlay, Evers, et~al.]{heidenreich20222022}
Paul~A Heidenreich, Biykem Bozkurt, David Aguilar, Larry~A Allen, Joni~J Byun,
  Monica~M Colvin, Anita Deswal, Mark~H Drazner, Shannon~M Dunlay, Linda~R
  Evers, et~al.
\newblock 2022 aha/acc/hfsa guideline for the management of heart failure: a
  report of the american college of cardiology/american heart association joint
  committee on clinical practice guidelines.
\newblock \emph{Journal of the American College of Cardiology}, 79\penalty0
  (17):\penalty0 e263--e421, 2022.

\bibitem[Hu et~al.(2014)Hu, Lu, and Tan]{hu2014discriminative}
Junlin Hu, Jiwen Lu, and Yap-Peng Tan.
\newblock Discriminative deep metric learning for face verification in the
  wild.
\newblock In \emph{Proceedings of the IEEE conference on computer vision and
  pattern recognition}, pages 1875--1882, 2014.

\bibitem[Huang et~al.(2018)Huang, Mu, Hu, Zou, and Xiao]{huang2018regression}
Zheng-Yi Huang, Qing-Qing Mu, Mian Hu, Ying Zou, and Jiang-Wen Xiao.
\newblock Regression-based metric learning.
\newblock In \emph{2018 37th Chinese Control Conference (CCC)}, pages
  9107--9112. IEEE, 2018.

\bibitem[Ilvento(2019)]{ilvento2019metric}
Christina Ilvento.
\newblock Metric learning for individual fairness.
\newblock \emph{arXiv preprint arXiv:1906.00250}, 2019.

\bibitem[Iscen et~al.(2018)Iscen, Tolias, Avrithis, and Chum]{iscen2018mining}
Ahmet Iscen, Giorgos Tolias, Yannis Avrithis, and Ond{\v{r}}ej Chum.
\newblock Mining on manifolds: Metric learning without labels.
\newblock In \emph{Proceedings of the IEEE Conference on Computer Vision and
  Pattern Recognition}, pages 7642--7651, 2018.

\bibitem[Jegou et~al.(2011)Jegou, Douze, and Schmid]{jegou2010product}
Herve Jegou, Matthijs Douze, and Cordelia Schmid.
\newblock Product quantization for nearest neighbor search.
\newblock \emph{IEEE transactions on pattern analysis and machine
  intelligence}, 33\penalty0 (1):\penalty0 117--128, 2011.

\bibitem[Kiyasseh et~al.(2021)Kiyasseh, Zhu, and Clifton]{kiyasseh2021clocs}
Dani Kiyasseh, Tingting Zhu, and David~A Clifton.
\newblock Clocs: Contrastive learning of cardiac signals across space, time,
  and patients.
\newblock In \emph{International Conference on Machine Learning}, pages
  5606--5615. PMLR, 2021.

\bibitem[Komajda and Lam(2014)]{komajda2014heart}
Michel Komajda and Carolyn~SP Lam.
\newblock Heart failure with preserved ejection fraction: a clinical dilemma.
\newblock \emph{European heart journal}, 35\penalty0 (16):\penalty0 1022--1032,
  2014.

\bibitem[Lan et~al.(2022)Lan, Ng, Hong, and Feng]{lan2022intra}
Xiang Lan, Dianwen Ng, Shenda Hong, and Mengling Feng.
\newblock Intra-inter subject self-supervised learning for multivariate cardiac
  signals.
\newblock In \emph{Proceedings of the AAAI Conference on Artificial
  Intelligence}, volume~36, pages 4532--4540, 2022.

\bibitem[Masetic and Subasi(2016)]{masetic2016congestive}
Zerina Masetic and Abdulhamit Subasi.
\newblock Congestive heart failure detection using random forest classifier.
\newblock \emph{Computer methods and programs in biomedicine}, 130:\penalty0
  54--64, 2016.

\bibitem[Mehari and Strodthoff(2022)]{mehari2022self}
Temesgen Mehari and Nils Strodthoff.
\newblock Self-supervised representation learning from 12-lead ecg data.
\newblock \emph{Computers in Biology and Medicine}, 141:\penalty0 105114, 2022.

\bibitem[Milbich et~al.(2021)Milbich, Roth, Sinha, Schmidt, Ghassemi, and
  Ommer]{milbich2021characterizing}
Timo Milbich, Karsten Roth, Samarth Sinha, Ludwig Schmidt, Marzyeh Ghassemi,
  and Bjorn Ommer.
\newblock Characterizing generalization under out-of-distribution shifts in
  deep metric learning.
\newblock \emph{Advances in Neural Information Processing Systems}, 34, 2021.

\bibitem[Morina et~al.(2019)Morina, Oliinyk, Waton, Marusic, and
  Georgatzis]{morina2019auditing}
Giulio Morina, Viktoriia Oliinyk, Julian Waton, Ines Marusic, and Konstantinos
  Georgatzis.
\newblock Auditing and achieving intersectional fairness in classification
  problems.
\newblock \emph{arXiv preprint arXiv:1911.01468}, 2019.

\bibitem[M{\"u}ller(2007)]{muller2007dynamic}
Meinard M{\"u}ller.
\newblock Dynamic time warping.
\newblock \emph{Information retrieval for music and motion}, pages 69--84,
  2007.

\bibitem[Nagueh et~al.(1997)Nagueh, Middleton, Kopelen, Zoghbi, and
  Qui{\~n}ones]{nagueh1997doppler}
Sherif~F Nagueh, Katherine~J Middleton, Helen~A Kopelen, William~A Zoghbi, and
  Miguel~A Qui{\~n}ones.
\newblock Doppler tissue imaging: a noninvasive technique for evaluation of
  left ventricular relaxation and estimation of filling pressures.
\newblock \emph{Journal of the American College of Cardiology}, 30\penalty0
  (6):\penalty0 1527--1533, 1997.

\bibitem[Nagueh et~al.(2009)Nagueh, Appleton, Gillebert, Marino, Oh, Smiseth,
  Waggoner, Flachskampf, Pellikka, and Evangelisa]{nagueh2009recommendations}
Sherif~F Nagueh, Christopher~P Appleton, Thierry~C Gillebert, Paolo~N Marino,
  Jae~K Oh, Otto~A Smiseth, Alan~D Waggoner, Frank~A Flachskampf, Patricia~A
  Pellikka, and Arturo Evangelisa.
\newblock Recommendations for the evaluation of left ventricular diastolic
  function by echocardiography.
\newblock \emph{European journal of echocardiography}, 10\penalty0
  (2):\penalty0 165--193, 2009.

\bibitem[Oh et~al.(2022)Oh, Chung, Kwon, Hong, and Choi]{oh2022lead}
Jungwoo Oh, Hyunseung Chung, Joon-myoung Kwon, Dong-gyun Hong, and Edward Choi.
\newblock Lead-agnostic self-supervised learning for local and global
  representations of electrocardiogram.
\newblock In \emph{Conference on Health, Inference, and Learning}, pages
  338--353. PMLR, 2022.

\bibitem[Paixao et~al.(2020)Paixao, Berriel, Boeres, Koerich, Badue, Souza, and
  Oliveira-Santos]{paixao2020fast}
Thiago~M Paixao, Rodrigo~F Berriel, Maria Boeres, Alessandro~L Koerich,
  Claudine Badue, Alberto F~De Souza, and Thiago Oliveira-Santos.
\newblock Fast (er) reconstruction of shredded text documents via
  self-supervised deep asymmetric metric learning.
\newblock In \emph{Proceedings of the IEEE/CVF Conference on Computer Vision
  and Pattern Recognition}, pages 14343--14351, 2020.

\bibitem[Pedregosa et~al.(2011)Pedregosa, Varoquaux, Gramfort, Michel, Thirion,
  Grisel, Blondel, Prettenhofer, Weiss, Dubourg, Vanderplas, Passos,
  Cournapeau, Brucher, Perrot, and Duchesnay]{scikit-learn}
F.~Pedregosa, G.~Varoquaux, A.~Gramfort, V.~Michel, B.~Thirion, O.~Grisel,
  M.~Blondel, P.~Prettenhofer, R.~Weiss, V.~Dubourg, J.~Vanderplas, A.~Passos,
  D.~Cournapeau, M.~Brucher, M.~Perrot, and E.~Duchesnay.
\newblock Scikit-learn: Machine learning in {P}ython.
\newblock \emph{Journal of Machine Learning Research}, 12:\penalty0 2825--2830,
  2011.

\bibitem[Raghu et~al.(2023{\natexlab{a}})Raghu, Chandak, Alam, Guttag, and
  Stultz]{raghu2023sequential}
Aniruddh Raghu, Payal Chandak, Ridwan Alam, John Guttag, and Collin Stultz.
\newblock Sequential multi-dimensional self-supervised learning for clinical
  time series.
\newblock In \emph{International Conference on Machine Learning}. PMLR,
  2023{\natexlab{a}}.

\bibitem[Raghu et~al.(2023{\natexlab{b}})Raghu, Schlesinger, Pomerantsev,
  Devireddy, Shah, Garasic, Guttag, and Stultz]{raghu2023ecg}
Aniruddh Raghu, Daphne Schlesinger, Eugene Pomerantsev, Srikanth Devireddy,
  Pinak Shah, Joseph Garasic, John Guttag, and Collin~M Stultz.
\newblock Ecg-guided non-invasive estimation of pulmonary congestion in
  patients with heart failure.
\newblock \emph{Scientific Reports}, 13\penalty0 (1):\penalty0 3923,
  2023{\natexlab{b}}.

\bibitem[Reyna et~al.(2021)Reyna, Sadr, Alday, Gu, Shah, Robichaux, Rad, Elola,
  Seyedi, Ansari, et~al.]{reyna2021will}
Matthew~A Reyna, Nadi Sadr, Erick A~Perez Alday, Annie Gu, Amit~J Shah, Chad
  Robichaux, Ali~Bahrami Rad, Andoni Elola, Salman Seyedi, Sardar Ansari,
  et~al.
\newblock Will two do? varying dimensions in electrocardiography: the
  physionet/computing in cardiology challenge 2021.
\newblock In \emph{2021 Computing in Cardiology (CinC)}, volume~48, pages 1--4.
  IEEE, 2021.

\bibitem[Roger(2013)]{roger2013epidemiology}
V{\'e}ronique~L Roger.
\newblock Epidemiology of heart failure.
\newblock \emph{Circulation research}, 113\penalty0 (6):\penalty0 646--659,
  2013.

\bibitem[Roth et~al.(2020)Roth, Milbich, Sinha, Gupta, Ommer, and
  Cohen]{roth2020revisiting}
Karsten Roth, Timo Milbich, Samarth Sinha, Prateek Gupta, Bjorn Ommer, and
  Joseph~Paul Cohen.
\newblock Revisiting training strategies and generalization performance in deep
  metric learning.
\newblock In \emph{International Conference on Machine Learning}, pages
  8242--8252. PMLR, 2020.

\bibitem[Roth et~al.(2021)Roth, Milbich, Ommer, Cohen, and
  Ghassemi]{roth2021simultaneous}
Karsten Roth, Timo Milbich, Bjorn Ommer, Joseph~Paul Cohen, and Marzyeh
  Ghassemi.
\newblock Simultaneous similarity-based self-distillation for deep metric
  learning.
\newblock In \emph{International Conference on Machine Learning}, pages
  9095--9106. PMLR, 2021.

\bibitem[Schlesinger et~al.(2022)Schlesinger, Diamant, Raghu, Reinertsen,
  Young, Batra, Pomerantsev, and Stultz]{schlesinger2022deep}
Daphne~E Schlesinger, Nathaniel Diamant, Aniruddh Raghu, Erik Reinertsen,
  Katherine Young, Puneet Batra, Eugene Pomerantsev, and Collin~M Stultz.
\newblock A deep learning model for inferring elevated pulmonary capillary
  wedge pressures from the 12-lead electrocardiogram.
\newblock \emph{JACC: Advances}, 1\penalty0 (1):\penalty0 100003, 2022.

\bibitem[Schroff et~al.(2015)Schroff, Kalenichenko, and
  Philbin]{schroff2015facenet}
Florian Schroff, Dmitry Kalenichenko, and James Philbin.
\newblock Facenet: A unified embedding for face recognition and clustering.
\newblock In \emph{Proceedings of the IEEE conference on computer vision and
  pattern recognition}, pages 815--823, 2015.

\bibitem[Sohn(2016)]{sohn2016improved}
Kihyuk Sohn.
\newblock Improved deep metric learning with multi-class n-pair loss objective.
\newblock \emph{Advances in neural information processing systems}, 29, 2016.

\bibitem[Tolia et~al.(2018)Tolia, Khan, Gholkar, and
  Zughaib]{tolia2018validating}
Sunit Tolia, Zubair Khan, Gunjan Gholkar, and Marcel Zughaib.
\newblock Validating left ventricular filling pressure measurements in patients
  with congestive heart failure: Cardiomems™ pulmonary arterial diastolic
  pressure versus left atrial pressure measurement by transthoracic
  echocardiography.
\newblock \emph{Cardiology Research and Practice}, 2018, 2018.

\bibitem[Wagner et~al.(2020)Wagner, Strodthoff, Bousseljot, Kreiseler, Lunze,
  Samek, and Schaeffter]{wagner2020ptb}
Patrick Wagner, Nils Strodthoff, Ralf-Dieter Bousseljot, Dieter Kreiseler,
  Fatima~I Lunze, Wojciech Samek, and Tobias Schaeffter.
\newblock Ptb-xl, a large publicly available electrocardiography dataset.
\newblock \emph{Scientific data}, 7\penalty0 (1):\penalty0 1--15, 2020.

\bibitem[Wang et~al.(2017)Wang, Zhou, Wen, Liu, and Lin]{wang2017deep}
Jian Wang, Feng Zhou, Shilei Wen, Xiao Liu, and Yuanqing Lin.
\newblock Deep metric learning with angular loss.
\newblock In \emph{Proceedings of the IEEE international conference on computer
  vision}, pages 2593--2601, 2017.

\bibitem[Waskom(2021)]{Waskom2021}
Michael~L. Waskom.
\newblock seaborn: statistical data visualization.
\newblock \emph{Journal of Open Source Software}, 6\penalty0 (60):\penalty0
  3021, 2021.
\newblock \doi{10.21105/joss.03021}.
\newblock URL \url{https://doi.org/10.21105/joss.03021}.

\bibitem[Wu et~al.(2017)Wu, Manmatha, Smola, and Krahenbuhl]{wu2017sampling}
Chao-Yuan Wu, R~Manmatha, Alexander~J Smola, and Philipp Krahenbuhl.
\newblock Sampling matters in deep embedding learning.
\newblock In \emph{Proceedings of the IEEE international conference on computer
  vision}, pages 2840--2848, 2017.

\bibitem[Yu et~al.(2018)Yu, Liu, Gong, Ding, and Tao]{yu2018correcting}
Baosheng Yu, Tongliang Liu, Mingming Gong, Changxing Ding, and Dacheng Tao.
\newblock Correcting the triplet selection bias for triplet loss.
\newblock In \emph{Proceedings of the European Conference on Computer Vision
  (ECCV)}, pages 71--87, 2018.

\bibitem[Yu et~al.(2021)Yu, Wang, Chen, and Guo]{yu2021automatic}
Junsheng Yu, Xiangqing Wang, Xiaodong Chen, and Jinglin Guo.
\newblock Automatic premature ventricular contraction detection using deep
  metric learning and knn.
\newblock \emph{Biosensors}, 11\penalty0 (3):\penalty0 69, 2021.

\bibitem[Zhang et~al.(2022)Zhang, Zhao, Tsiligkaridis, and
  Zitnik]{zhang2022self}
Xiang Zhang, Ziyuan Zhao, Theodoros Tsiligkaridis, and Marinka Zitnik.
\newblock Self-supervised contrastive pre-training for time series via
  time-frequency consistency.
\newblock \emph{Advances in Neural Information Processing Systems},
  35:\penalty0 3988--4003, 2022.

\bibitem[Zheng et~al.(2020{\natexlab{a}})Zheng, Chu, Struppa, Zhang, Yacoub,
  El-Askary, Chang, Ehwerhemuepha, Abudayyeh, Barrett,
  et~al.]{zheng2020optimal}
Jianwei Zheng, Huimin Chu, Daniele Struppa, Jianming Zhang, Magdi Yacoub,
  Hesham El-Askary, Anthony Chang, Louis Ehwerhemuepha, Islam Abudayyeh,
  Alexander Barrett, et~al.
\newblock Optimal multi-stage arrhythmia classification approach.
\newblock \emph{Scientific reports}, 10\penalty0 (1):\penalty0 1--17,
  2020{\natexlab{a}}.

\bibitem[Zheng et~al.(2020{\natexlab{b}})Zheng, Zhang, Danioko, Yao, Guo, and
  Rakovski]{zheng202012}
Jianwei Zheng, Jianming Zhang, Sidy Danioko, Hai Yao, Hangyuan Guo, and Cyril
  Rakovski.
\newblock A 12-lead electrocardiogram database for arrhythmia research covering
  more than 10,000 patients.
\newblock \emph{Scientific data}, 7\penalty0 (1):\penalty0 1--8,
  2020{\natexlab{b}}.

\bibitem[Zhu et~al.(2022)Zhu, Ma, Huang, Wang, and Yang]{zhu2022dual}
Guiping Zhu, Mingzhu Ma, Yuwen Huang, Kuikui Wang, and Gongping Yang.
\newblock Dual-domain low-rank fusion deep metric learning for off-the-person
  ecg biometrics.
\newblock In \emph{ICASSP 2022-2022 IEEE International Conference on Acoustics,
  Speech and Signal Processing (ICASSP)}, pages 2914--2918. IEEE, 2022.

\end{thebibliography}
\clearpage
\appendix
\section*{Appendix}
\label{appendix}
In the appendix section of this work, we provide a more detailed and formalized explanation of various DML objectives, batch mining strategies, and loss functions introduced in the main paper. Specifically, we introduce the formalization for supervised DML joint-learning classification and regression objectives, as well as self-supervised downstream objectives (Sections \ref{appendix_supdml}). Additionally, we introduce other loss objectives, including margin loss and angular loss (Section \ref{dml_objectives}) and discuss mining techniques incorporating slack parameters into hard mining objectives, such as semi-hard and soft-hard mining (Section \ref{mining_appendix}). We then introduce the results and discussions for the subgroup performance gap (Section \ref{subgroup_appendix} and explain clinical implications (Section \ref{appendix_clinicalimplication}).

Throughout the Appendix, we used the same notation introduced in Section \ref{methods}: $x_i$ and $y_i$ are input data instances and labels, respectively, in $D_{sup}$ with the cardinal $N$. Instances in $D_{sup}$ are sampled for triplet mining, where $x_a$ is an anchor, $x_p$ is positive, and $x_n$ is negative. $f_\Phi$ is the metric embedding where we apply the DML objectives. For either joint training or finetuning hemodynamics inference, we use additional FC layer $g$.

\section{Supervised DML and Self-supervised DML Learning Objectives}
\label{appendix_supdml}
\subsection{Supervised Deep Metric Learning for Classification}
\label{supdml_class}

Below is the detailed formulation of the DML objective of SupDML joint classification, where $\alpha$ is the scaling coefficient of triplet loss and $\sigma$ is the sigmoid function. 

\begin{align*}
    \mathcal{L}_{tot} &= \mathcal{L}_{CE} + \alpha\mathcal{L}_{triplet} \nonumber\\
    &= -\sum_{i=1}^N\ y_i log(\sigma(g(f_\Phi(x_i))) + \alpha\sum_{i=1}^N \{||f_\Phi(x_a)-f_\Phi(x_p)|| - ||f_\Phi(x_a)-f_\Phi(x_n)||\} 
\end{align*}

\begin{table*}[t!]
\footnotesize
 \centering
  \caption{Performance of downstream mPCWP classification task with various triplet loss weight coefficient $\alpha$. We report the mean performance and its standard deviation of 1,0000 bootstraps. Also, the optimal performance for each dimension is highlighted in bold.}
 {
 	\begin{tabular}{c|c|ccc}
 		\toprule
  		& & \multicolumn{2}{c}{Classification} &\multicolumn{1}{c}{Regression}\\
 		\cmidrule(r){2-2}\cmidrule(r){3-4}\cmidrule(r){5-5}
 		Dim & $\alpha$ & AUC & APR & RMSE  \\
            \midrule
		\multirow{5}{*}{128}
		& 0.1 
            & 75.4 $\pm$ 1.8
 		& \textbf{57.2 $\pm$ 3.3}
 		& 7.37 $\pm$ 0.25
 		\\
 		& 1.0 
            & 74.6 $\pm$ 1.8
 		& \textbf{57.2 $\pm$ 3.2}
 		& 7.50 $\pm$ 0.22 
 		\\
  		& 2.0 
            & 74.6 $\pm$ 1.8
 		& 56.2 $\pm$ 3.3
 		& \textbf{7.15 $\pm$ 0.21} 
  		\\
            & 3.0 
            & \textbf{76.7 $\pm$ 1.6} 
 		& 56.1 $\pm$ 3.3
 		& 7.23 $\pm$ 0.21 
 		\\
            & 10.0
            & 76.2 $\pm$ 1.7 
 		& 55.0 $\pm$ 3.1
 		& 7.27 $\pm$ 0.20 
 		\\
 		\midrule
		\multirow{5}{*}{256}
		& 0.1
            & 73.1 $\pm$ 1.8 
 		& 54.2 $\pm$ 3.2
  		& 7.53 $\pm$ 0.22 
 		\\
 		& 1.0
            & 75.3 $\pm$ 1.7 
 		& 53.2 $\pm$ 3.1
 		& 7.55 $\pm$ 0.21 
 		\\
            & 2.0
            & \textbf{76.5 $\pm$ 1.7} 
 		& \textbf{57.2 $\pm$ 3.2}
 		& 7.27 $\pm$ 0.21 
  		\\
            & 3.0
            & 75.5 $\pm$ 1.7 
 		& 57.1 $\pm$ 3.2
 		& \textbf{7.19 $\pm$ 0.20}
 		\\
            & 10.0
            & 74.1 $\pm$ 1.7
 		& 53.4 $\pm$ 3.1
 		& 7.37 $\pm$ 0.22 
 		\\
 		\midrule
		\multirow{5}{*}{512}
		& 0.1 
            & 75.8 $\pm$ 1.8 
 		& 55.2 $\pm$ 3.3
 		& 7.67 $\pm$ 0.24 
 		\\
 		& 1.0 
            & \textbf{77.1 $\pm$ 1.8}
 		& \textbf{59.6 $\pm$ 3.2}
 		& 7.44 $\pm$ 0.23 
 		\\
  		& 2.0 
            & 76.6 $\pm$ 1.6
 		& 55.8 $\pm$ 3.2
 		& 7.30 $\pm$ 0.23 
  		\\
            & 3.0 
            & 74.9 $\pm$ 1.7
 		& 56.9 $\pm$ 3.2
 		& \textbf{7.21 $\pm$ 0.18}
 		\\
            & 10.0 
            & 75.6 $\pm$ 1.7
 		& 55.5 $\pm$ 3.1
 		& 7.32 $\pm$ 0.22
 		\\
 		\midrule
		\multirow{5}{*}{1024}
		& 0.1 
            & 74.9 $\pm$ 1.7
 		& 55.8 $\pm$ 3.2
 		& 7.66 $\pm$ 0.23 
 		\\
 		& 1.0 
            & \textbf{77.5 $\pm$ 1.7}
 		& \textbf{58.1 $\pm$ 2.9}
 		& 7.42 $\pm$ 0.18
 		\\
  		& 2.0 
            & 76.0 $\pm$ 1.7
 		& 55.7 $\pm$ 3.3
 		& 7.43 $\pm$ 0.20
  		\\
            & 3.0 
            & 75.0 $\pm$ 1.8
 		& 54.9 $\pm$ 3.3
 		& \textbf{7.20 $\pm$ 0.20} 
 		\\
            & 10.0 
            & 76.1 $\pm$ 1.7
 		& 54.8 $\pm$ 3.1
 		& 7.32 $\pm$ 0.21
 		\\
            \bottomrule
 	\end{tabular}
 	}
 \label{alpha}
\end{table*}

\subsection{Supervised Deep Metric Learning for Regression}
\label{supdml_reg}
Using the triplet $(x_a, x_p, x_n)$, the model $f_\Phi$ is optimized by triplet loss ($\mathcal{L}_{Triplet}$) while at the same time the final classifier $g \cdot f_\Phi$ is jointly optimized with Root Mean Square Error (RMSE) with the triplet scaling factor of $\alpha$: $\mathcal{L}_{tot} = \mathcal{L}_{RMSE} + \alpha\mathcal{L}_{triplet}$. RMSE for downstream classification loss is $\mathcal{L}_{RMSE} = \sqrt{\frac{\sum\limits_{i=1}^{N}\vert\vert y_i - g(f_\Phi(x_i))\vert\vert^2}{n}}$ and the equation for the full loss term is:

\begin{align*}
    \mathcal{L}_{tot} &= \mathcal{L}_{RMSE} + \alpha\mathcal{L}_{triplet} \\
    &= \sqrt{\frac{\sum\limits_{i=1}^{N}\vert\vert y_i - g(f_\Phi(x_i))\vert\vert^2}{n}} + \alpha\sum_{i=1}^N \{||f_\Phi(x_a)-f_\Phi(x_p)|| - ||f_\Phi(x_a)-f_\Phi(x_n)||\} 
\end{align*}

\subsection{Optimal Scaling Coefficient $\alpha$ for SupDML Classification and Regression}
\label{optimal_alpha}
We present the results of our search for the optimal scaling factor $\alpha$ for the triplet loss and its effects on performance. We performed a greedy search for the best scaling factor $\alpha$ among the values 0.1, 1.0, 2.0, 3.0, and 10.0 for each embedding dimension of $f_\Phi$ (128, 256, 512, 1024) (Table \ref{alpha}).

We report the best performance among different $\alpha$ values for the main supDML results (Table \ref{result}): $\alpha = 3.0$ yielded the best performance for the classification of the supDML experiment with an embedding $f_\Phi$ dimension of 128, while $\alpha = 2.0$ was optimal for joint regression with a 128-dimensional embedding. For an embedding $f_\Phi$ dimension of 256, $\alpha = 2.0$ was best for joint classification, and $\alpha = 3.0$ was best for regression. With an embedding $f_\Phi$ dimension of 512, $\alpha = 1.0$ worked best for joint classification, and $\alpha = 3.0$ was optimal for regression. Lastly, for an embedding $f_\Phi$ dimension of 1024, $\alpha = 1.0$ was best for joint classification, and $\alpha = 3.0$ was optimal for regression.

\subsection{Downstream Hemodynamics Inference for Self-Supervised DML}
\label{finetuning}
The pre-trained model $f_\Phi$ with the DML objective and finetune with the downstream hemodynamics classification and regression task with $\mathcal{D}_{sup}$. The objective uses the cross-entropy loss for binary classification ($\mathcal{L}_{CE}$) and RMSE for regression ($\mathcal{L}_{RMSE}$):

\begin{align*}
    \mathcal{L}_{CE}=-\sum_{i=1}^N\ y_i log(\sigma(f_\Phi(x_i))\\
    \mathcal{L}_{RMSE} = \sqrt{\frac{\sum\limits_{i=1}^{N}\vert\vert y_i - g(f_\Phi(x_i))\vert\vert^2}{n}}
\end{align*}

\section{Deep Metric Learning Loss Functions}
\label{dml_objectives}

We used two different DML objectives to extend the use of triplet loss: margin loss and angular loss.

\paragraph{Margin Loss} \citep{wu2017sampling} extends the standard triplet loss by incorporating a dynamic, learnable distance boundary $\beta$ between the samples with the same labels and different labels (relative ordering, $\mathcal{P} =\{(x_i, x_j)|x_i \in \mathcal{B}, x_j \in \mathcal{B}\}$), which has transformed from the usual triplet ranking problem. The learning rate of the boundary, $\beta$ is set to 0.0005, with an initial $\beta$ value of 1.2 and a triplet margin $\gamma$ of 0.2. The implementation and hyperparameters followed the code from \cite{roth2020revisiting}.

\begin{align*}
   \mathcal{L}_{margin} = \sum_{(x_i, x_j) \in \mathcal{P}} \mathbb{1}_{y_i=y_j} (||f_\Phi(x_i)-f_\Phi(x_j)||_2^2 -\beta) - \mathbb{1}_{y_i \neq y_j} (||f_\Phi(x_i)-f_\Phi(x_j)||_2^2 -\beta)+ \gamma
\end{align*}

\paragraph{Angular Loss} \citep{wang2017deep} The angular loss constrains the angle at the negative point of triplet triangles, resulting in higher convergence, achieving scale invariance and third-order geometric restrictions. We used the proposed parameter from the original paper, angular margin $\alpha = \frac{\pi}{4}$, and the parameter for the trade-off between npair and angular loss, $\lambda=2$. The objective function below uses the triplets anchor $x_a$, positive $x_p$ and negative $x_n$:
{
\small
\begin{align*}
   \mathcal{L}_{angular} = \mathcal{L}_{npair} + \frac{\lambda}{b} \sum [\log(1+\sum \exp(4 \tan^2 (\alpha (f_\Phi(x_a)+f_\Phi(x_p))^T f_\Phi(x_n)))-2(1+\tan^2(\alpha))f_\Phi(x_a)^Tf_\Phi(x_n)]
\end{align*}
}
where $\mathcal{L}_{npair}$ is defined as below with the embedding regularization $\nu = 0.005$ following the original paper \citep{sohn2016improved}:
\begin{align*}
    \mathcal{L}_{npair} = \frac{1}{b} \sum \log(1+\sum \exp (f_\Phi(x_a)^Tf_\Phi(x_n) - f_\Phi(x_a)^Tf_\Phi(x_p))) + \frac{\nu}{b} \sum_{x_i \in \mathcal{B}} ||f_\Phi(x_i)||^2_2
\end{align*}

Table \ref{loss} summarizes the classification and regression performance with margin loss and angular loss, compared with triplet loss.

\begin{table*}[h!]
\footnotesize
 \centering
  \caption{Performance of the downstream Wedge pressure elevation classification task with various loss functions for supervised DML with an embedding dimension of 128.} 
 {
 	\begin{tabular}{c|ccc}
 		\toprule
  		& \multicolumn{2}{c}{Classification} &\multicolumn{1}{c}{Regression}\\
 		\cmidrule(r){2-3}\cmidrule(r){4-4}
 		& AUC & APR & RMSE  \\
 		\midrule
		Triplet Loss
            & \textbf{76.7 $\pm$ 1.6}
 		& \textbf{56.1 $\pm$ 3.3}
 		& \textbf{7.15} $\pm$ 0.21 
 		\\
            \midrule
 		Margin Loss
 		& 74.7 $\pm$ 1.7
 		& 53.4 $\pm$ 3.3
 		& 7.55 $\pm$ 0.22
 		\\
            \midrule
  		Angular Loss
 		& \textbf{76.8 $\pm$ 1.8}
 		& \textbf{56.8 $\pm$ 3.1}
 		& 7.59 $\pm$ 0.22
  		\\
 		\bottomrule
 	\end{tabular}
 	}
 \label{loss}
\end{table*}

\section{Triplet Mining for Supervised DML}
\label{mining_appendix}
We formalize various mining techniques we used for supervised DML, including continuous label-based mining (\textit{Label Based}) and hard negative mining techniques (\textit{Semihard}, \textit{Softhard}).

\subsection{Continuous Label Based Mining}
\label{labelbased}
\paragraph{DML Triplet Selection:} For a given anchor $x_a$, we define the positive sample $x_p$ to be the input sample with the smallest absolute label difference with the anchor: $x_p = \{x_i|i = \underset{y_i \in \mathcal{B}}{\arg \min }|y_{i} - y_{a}|\}$ where $\mathcal{B}$ is the batch. Then the negative sample $x_n$ is then randomly sampled from the set $\mathcal{N}_\mathcal{B} = \{x_j|j = \underset{y_j \in \mathcal{B}}{\arg \max } |y_{j} - y_{a}|\}$. We also performed greedy search hyperparameter tuning for the scaling factor $\alpha$ for label based mining loss and reported the best-performing model in Table \ref{mining}. With an embedding dimension $f_\Phi$ of dimensions 128 and 256, $\alpha$ 10.0 worked the best, and for 512 and 1024 $\alpha$ of 0.1 worked the best.

\begin{table*}[h!]
\footnotesize
 \centering
  \caption{Performance of the downstream mPCWP elevation classification and regression task with various mining methods for supervised DML (Random, Semihard, Softhard)} 
 {
 	\begin{tabular}{c|c|cccc}
 		\toprule
  	    & & \multicolumn{2}{c}{Classification} &\multicolumn{1}{c}{Regression} &\multicolumn{1}{c}{Embedding}\\
 		\cmidrule(r){1-1}\cmidrule(r){2-2}\cmidrule(r){3-4}\cmidrule(r){5-5}\cmidrule(r){6-6}
            Mining & Dim & AUC & APR & RMSE  & Recall@1 \\
            \midrule
		\multirow{4}{*}{Random} 
		& 128 
            & 76.7 $\pm$ 1.6
 		& 56.1 $\pm$ 3.3
 		& \textbf{7.15} $\pm$ 0.21
            & \textbf{69.6}
		\\
 		& 256
            & 76.5 $\pm$ 1.7
 		& 57.2 $\pm$ 3.2
 		& 7.19 $\pm$ 0.20
            & 66.5
 		\\
  		& 512
            & \textbf{77.1 $\pm$ 1.8}
 		& \textbf{59.6 $\pm$ 3.2}
 		& 7.21 $\pm$ 0.18
            & 67.5
  		\\
            & 1024 
            & \textbf{77.5 $\pm$ 1.7}
 		& 58.1 $\pm$ 2.9
 		& 7.20 $\pm$ 0.20
            & 66.3
 		\\
 		\midrule
		\multirow{4}{*}{Label Based} 
		& 128 
            & 76.1 $\pm$ 1.7
 		& 56.5 $\pm$ 1.3
 		& 7.29 $\pm$ 0.21
            & 67.5
		\\
 		& 256
            & 75.8 $\pm$ 1.8
 		& 55.6 $\pm$ 3.2
 		& 7.39 $\pm$ 0.20
            & 67.4
 		\\
  		& 512
            & 74.9 $\pm$ 1.8
 		& 54.8 $\pm$ 3.3
 		& 7.40 $\pm$ 0.24
            & 59.1
  		\\
            & 1024 
            & \textbf{77.4 $\pm$ 1.6}
 		& 57.6 $\pm$ 3.1
 		& 7.37 $\pm$ 0.23
            & 65.4
 		\\
 		\midrule
		\multirow{4}{*}{Semihard}
		& 128
 		& 75.1 $\pm$ 1.8
 		& 56.4 $\pm$ 3.4
 		& 7.50 $\pm$ 0.21
            & 66.2
 		\\
 		& 256
 		& 76.9 $\pm$ 1.7
 		& 56.5 $\pm$ 3.4
 		& 7.51 $\pm$ 0.20
            & 66.2
 		\\
  		& 512
 		& 75.0 $\pm$ 1.8
 		& 57.5 $\pm$ 3.1
 		& 7.46 $\pm$ 0.22
            & 65.3
  		\\
            & 1024
 		& 74.4 $\pm$ 1.7
 		& 51.5 $\pm$ 3.2
 		& \textbf{7.19 $\pm$ 0.21}
            & 66.1
 		\\
 		\midrule
            \multirow{4}{*}{Softhard}
		& 128    
 		& \textbf{77.4 $\pm$ 1.7}
 		& 58.1 $\pm$ 3.1
 		& 7.53 $\pm$ 0.22
            & 67.9
 		\\
 		& 256
 		& 75.9 $\pm$ 1.7 
 		& 57.7 $\pm$ 3.2
 		& 7.38 $\pm$ 0.21
            & \textbf{69.3}
 		\\
  		& 512
            & 74.3 $\pm$ 1.7
 		& 54.0 $\pm$ 3.1
 		& 7.50 $\pm$ 0.21
            & 67.9
  		\\
            & 1024
 		& 74.8 $\pm$ 1.8
 		& 52.2 $\pm$ 3.3
            & 7.51 $\pm$ 0.21
            & 65.4
 		\\
 		\bottomrule
 	\end{tabular}
 	}
 \label{mining}
\end{table*}

\subsection{Hard Negative Mining}
\label{hardnegative}
\paragraph{Semihard Mining} \citep{schroff2015facenet} is a method to sample hard negatives based on the distance of negative samples with respect to anchors and positives. Semihard mining selects the fairly hard negatives with the embedding distance to the anchor embedding larger than that is from positive embedding, within a certain distance margin. If we let the embedding of negative samples that are to be selected for triplet pair $f_\Phi(x_n)$, then the selected negatives are:
\begin{align*}
    x_n \in \{x_n|x_n \in \mathcal{B}, y_n \neq y_a, ||f_\Phi(x_a)-f_\Phi(x_p)||_2^2 < ||f_\Phi(x_a)-f_\Phi(x_n)||_2^2\}
\end{align*}

\paragraph{Softhard Mining} \citep{roth2020revisiting} introduces a soft (probabilistic) selection of hard negatives. As observed by \cite{roth2020revisiting}, this decreases the likelihood of potential model collapses and undesirable local minima, which were the issues with semihard mining \citep{schroff2015facenet}. We select the hard negatives based on the two equations below:
\begin{align*}
    x_n \in \{x_n|x_n \in \mathcal{B}, y_n \neq y_a, ||f_\Phi(x_a)-f_\Phi(x_n)||_2^2 < \argmax_{x_p \in \mathcal{B}, y_a= y_p}||f_\Phi(x_a)-f_\Phi(x_p)||_2^2\}
\end{align*}
and
\begin{align*}
    x_n \in \{x_n|x_n \in \mathcal{B}, y_n \neq y_a, ||f_\Phi(x_a)-f_\Phi(x_n)||_2^2 > \argmin_{x_n \in \mathcal{B}, y_a \neq y_n}||f_\Phi(x_a)-f_\Phi(x_n)||_2^2\}
\end{align*}

Table \ref{mining} summarizes the classification and regression performance with hard negative minings compared with random mining. Table \ref{mining_subgroup} reports the performance of each gender subgroup and the performance gap. 
\begin{table*}[h!]
\footnotesize
 \centering
  \caption{Subgroup performance for the downstream Wedge pressure elevation classification task with various mining methods for supervised DML (Random, Semihard, Softhard). The embedding dimension used here is 128.} 
 {
 	\begin{tabular}{c|c|ccc}
 		\toprule
  		& Test & \multicolumn{2}{c}{Classification} &\multicolumn{1}{c}{Regression}\\
 		\cmidrule(r){2-2}\cmidrule(r){3-4}\cmidrule(r){5-5}
 		Mining & Subgroup & AUC & APR & RMSE  \\
            \midrule
		\multirow{4}{*}{Random}  
 		& Full
            & 76.7 $\pm$ 1.6
 		& 56.1 $\pm$ 3.3
 		& 7.15 $\pm$ 0.21 
 		\\
            & Male
  		& 77.4 $\pm$ 1.5
 		& 54.2 $\pm$ 2.8
 		& 7.07 $\pm$ 0.19
 		\\
  		& Female
 		& 75.6 $\pm$ 1.2
 		& 56.4 $\pm$ 2.4
 		& 7.34 $\pm$ 0.17
  		\\
  		& Gap (M-F)
 		& 1.8 
 		& 2.2
 		& 0.27 
            \\
 		\midrule
   		\multirow{4}{*}{Label Based}  
 		& Full
            & 76.1 $\pm$ 1.7
 		& 56.5 $\pm$ 1.3
 		& 7.29 $\pm$ 0.21
 		\\
            & Male
  		& 75.3 $\pm$ 1.0
 		& 54.7 $\pm$ 1.9
 		& 7.11 $\pm$ 0.17
 		\\
  		& Female
 		& 75.4 $\pm$ 1.3
 		& 57.3 $\pm$ 2.5
 		& 7.40 $\pm$ 0.17
  		\\
  		& Gap (M-F)
 		& \textbf{0.1}
 		& 2.5
 		& 0.29
            \\
 		\midrule
		\multirow{4}{*}{Semihard}
		& Full
 		& 75.1 $\pm$ 1.8 
 		& 56.4 $\pm$ 3.4
 		& 7.50 $\pm$ 0.21
 		\\
 		& Male
 		& 77.5 $\pm$ 1.6
 		& 56.3 $\pm$ 3.0
 		& 7.39 $\pm$ 0.18
 		\\
  		& Female
 		& 73.9 $\pm$ 1.4
 		& 56.9 $\pm$ 2.3
 		& 7.49 $\pm$ 0.26
  		\\
  		& Gap (M-F)
 		& 3.6
 		& \textbf{0.5}
 		& \textbf{0.10}
  		\\
 		\midrule
            \multirow{4}{*}{Softhard} 
		& Full
 		& 77.4 $\pm$ 1.7
 		& 58.1 $\pm$ 3.1
 		& 7.53 $\pm$ 0.22
 		\\
 		& Male
 		& 78.5 $\pm$ 1.5
 		& 54.5 $\pm$ 3.0
 		& 7.71 $\pm$ 0.18
 		\\
            & Female
            & 74.9 $\pm$ 1.3
 		& 57.0 $\pm$ 2.4
 		& 7.92 $\pm$ 0.17
 		\\
  		& Gap (M-F)
 		& 3.6
 		& 2.5
 		& 0.21
  		\\
 		\bottomrule
 	\end{tabular}
 	}
 \label{mining_subgroup}
\end{table*}

\section{Subgroup Performance Gap}
\label{subgroup_appendix}

In this paper, we showed the biases in data influencing the model performance, which can be improved using a supervised and self-supervised DML training scheme. Our dataset is over-represented by certain demographic groups (gender, age), and models could potentially perform better on these groups than on under-represented groups. As the female demographic group and age group who are between 18 to 35 are under-represented in our training data, baselines may be less accurate in predicting outcomes for these groups. This is why it is crucial to perform fairness audits and continue to evaluate the deployed performance of such models in other tasks. 

\subsection{Prevalence and Performance Metrics}
Among our performance metrics, APR could be dependent on prevalence, as it influences the precision values. However, we found that the prevalence of the positive class  (mPCWP $>$ 18) across gender groups in our dataset is similar: Female/Male prevalence ratio of 0.461/0.441 for the Train set, 0.395/0.410 for the Validation set, and 0.466/0.385 for the Test set. The APR is the correct metric in this case because it provides a measure of model performance across various threshold levels and is less sensitive to class imbalances than other metrics such as accuracy. 

\subsection{Age Subgroup Performance Gap}
\label{appendix_agegap}
We present a summary of performance gaps in mPCWP inference across age groups in Table \ref{agegap}. The gaps in performance are the averaged relative pairwise performance differences between all combinations of age groups. Both DML models achieved fairer downstream task performance across different age subgroups against the baselines. Compared to the supervised baseline, supervised DML not only achieved narrower gaps in both AUC and APR but also surpassed overall regression performance, while maintaining equivalent classification performance. Self-supervised DML showed lower APR and RMSE gaps compared to random initialization and self-supervised contrastive learning baselines (SimCLR, CLOCS, PCLR).

\begin{table*}[t!]
\tiny
 \centering
  \caption{Age subgroup performance gap of downstream mPCWP classification task has decreased on either Supervised or Self-Supervised DML compared to supervised and contrastive baselines (with an embedding dimension of 128 for DML and contrastive learning encoder). We report the mean with a standard deviation based on 1,000 bootstraps, and models with the lowest performance gaps are boldfaced. We included four age subgroups: 18-35, 35-50, 50-75, 75-. The average gap indicates the average pairwise group performance gap}
 {
 	\begin{tabular}{cccccc}
 		\toprule
  		& & & \multicolumn{2}{c}{Classification} &\multicolumn{1}{c}{Regression}\\
 		\cmidrule(r){4-5}\cmidrule(r){6-6}
 		& Models & Subgroup & AUC & APR & RMSE  \\
 		\midrule
            \multicolumn{2}{c}{\multirow{6}{*}{Random Init}}
 		& Full
            & 74.3 $\pm$ 1.9
 		& 55.7 $\pm$ 3.3
 		& 7.88 $\pm$ 0.26
 		\\
            &
     	& 18-35
 		& 80.6 $\pm$ 12.8
 		& 53.0 $\pm$ 21.9
 		& 3.63 $\pm$ 0.52
  		\\
            &
  		& 35-50
 		& 74.7 $\pm$ 1.8
 		& 54.4 $\pm$ 1.8
 		& 7.99 $\pm$ 0.76
  		\\
            &
  		& 50-75
 		& 74.6 $\pm$ 0.9
 		& 53.8 $\pm$ 1.6
 		& 8.51 $\pm$ 0.55
  		\\
            &
  		& 75-
 		& 74.6 $\pm$ 0.8
 		& 54.8 $\pm$ 1.6
 		& 8.64 $\pm$ 0.82
  		\\
            &
  		& Average Gap
 		& 3.0
 		& 1.0
 		& 2.6
  		\\
            \midrule
 		\multirow{12}{*}{Supervised}
            & \multirow{6}{*}{Supervised Baseline}
 		& Full
 		& 78.4 $\pm$ 1.1
 		& 59.5 $\pm$ 2.2
 		& 7.45 $\pm$ 0.15
 		\\
            & 
  		& 18-35
 		& 83.5 $\pm$ 5.4
 		& 59.4 $\pm$ 11.7
 		& 5.07 $\pm$ 0.51
  		\\
            &
  		& 35-50
 		& 73.2 $\pm$ 4.2
 		& 59.5 $\pm$ 6.3
 		& 7.41 $\pm$ 0.46
  		\\
            &
  		& 50-75
 		& 79.8 $\pm$ 1.5
 		& 55.6 $\pm$ 3.1
 		& 7.30 $\pm$ 0.18
  		\\
            &
  		& 75-
 		& 66.7 $\pm$ 2.8
 		& 69.1 $\pm$ 3.6
 		& 8.28 $\pm$ 0.37
  		\\
            &
  		& Average Gap
 		& 9.5
 		& 6.8
 		& \textbf{1.62}
  		\\
            \cmidrule{2-6}
		& \multirow{6}{*}{\begin{tabular}[c]{@{}c@{}}Supervised DML\\ (Random Mining)\end{tabular}}
 		& Full
		& 77.3 $\pm$ 3.6
 		& 57.8 $\pm$ 3.3
 		& 7.33 $\pm$ 0.23
 		\\
            & 
  		& 18-35
 		& 77.4 $\pm$ 1.6
 		& 57.5 $\pm$ 3.3
 		& 3.07 $\pm$ 0.39
  		\\
            &
  		& 35-50
 		& 77.3 $\pm$ 1.6
 		& 56.6 $\pm$ 3.1
 		& 7.31 $\pm$ 0.23
  		\\
            &
  		& 50-75
 		& 77.8 $\pm$ 2.2
 		& 60.0 $\pm$ 3.9
 		& 6.94 $\pm$ 0.21
  		\\
            &
  		& 75-
 		& 77.4 $\pm$ 1.2
 		& 57.6 $\pm$ 2.4
 		& 7.33 $\pm$ 0.16
  		\\
            &
  		& Average Gap
 		& \textbf{0.3}
 		& \textbf{1.7}
 		& 2.19
  		\\
            \midrule
            \midrule
            \multirow{16}{*}{Self-Supervised}
 		& \multirow{6}{*}{SimCLR}
 		& Full
 		& 75.0 $\pm$ 1.9
 		& 58.2 $\pm$ 3.2
 		& 7.58 $\pm$ 0.22
 		\\
            & 
  		& 18-35
 		& 74.8 $\pm$ 1.7
 		& 57.1 $\pm$ 3.1
 		& 3.93 $\pm$ 0.64
  		\\
            &
  		& 35-50
 		& 74.2 $\pm$ 1.7
 		& 55.8 $\pm$ 3.0
 		& 7.32 $\pm$ 0.84
  		\\
            &
  		& 50-75
 		& 74.8 $\pm$ 2.3
 		& 52.3 $\pm$ 4.1
 		& 7.32 $\pm$ 0.22
  		\\
            &
  		& 75-
 		& 74.7 $\pm$ 1.3
 		& 56.7 $\pm$ 2.2
 		& 7.41 $\pm$ 0.15
  		\\
            &
  		& Average Gap
 		& \textbf{0.3}
 		& 2.5
 		& 1.74
  		\\
            \cmidrule{2-6}            
 		& \multirow{6}{*}{CLOCS}
 		& Full
		& 75.9 $\pm$ 1.7
 		& 56.2 $\pm$ 3.4
 		& 7.45 $\pm$ 0.22
 		\\
            & 
  		& 18-35
 		& 73.5 $\pm$ 1.9
 		& 52.9 $\pm$ 3.2
 		& 4.47 $\pm$ 0.89
  		\\
            &
  		& 35-50
 		& 72.7 $\pm$ 1.7
 		& 52.5 $\pm$ 2.9
 		& 7.50 $\pm$ 0.70
  		\\
            &
  		& 50-75
 		& 74.5 $\pm$ 2.3
 		& 51.5 $\pm$ 3.9
 		& 7.50 $\pm$ 0.26
  		\\
            &
  		& 75-
 		& 73.4 $\pm$ 1.3
 		& 53.6 $\pm$ 2.2
 		& 7.66 $\pm$ 0.17
  		\\
            &
  		& Average Gap
 		& 0.9
 		& 1.1
 		& 1.60
  		\\
            \cmidrule{2-6}
 		& \multirow{6}{*}{PCLR}
 		& Full
		& 74.2 $\pm$ 1.6
 		& 54.2 $\pm$ 2.5
 		& 10.04 $\pm$ 0.27
 		\\
            & 
  		& 18-35
 		& 65.6 $\pm$ 16.1
 		& 65.9 $\pm$ 15.3
 		& 8.34 $\pm$ 1.31
  		\\
            &
  		& 35-50
 		& 73.1 $\pm$ 4.6
 		& 57.2 $\pm$ 7.6
 		& 11.29 $\pm$ 0.84
  		\\
            &
  		& 50-75
 		& 74.9 $\pm$ 1.9
 		& 51.7 $\pm$ 3.3
 		& 10.58 $\pm$ 0.30
  		\\
            &
  		& 75-
 		& 63.8 $\pm$ 3.8
 		& 61.1 $\pm$ 4.7
 		& 11.72 $\pm$ 0.69
  		\\
            &
  		& Average Gap
 		& 6.8
 		& 15.7
 		& 1.81
  		\\
            \cmidrule{2-6}            
 		& \multirow{6}{*}{\begin{tabular}[c]{@{}c@{}}Self-supervised DML\\ (DTW)\end{tabular}}
 		& Full
            & 77.2 $\pm$ 1.7
 		& 60.5 $\pm$ 3.1
 		& 7.65 $\pm$ 0.21
 		\\
            & 
  		& 18-35
            & 74.9 $\pm$ 1.0
 		& 54.7 $\pm$ 1.8
 		& 6.97 $\pm$ 1.22
  		\\
            &
  		& 35-50
 		& 74.8 $\pm$ 0.9
 		& 54.6 $\pm$ 1.8
 		& 7.68 $\pm$ 0.74
  		\\
            &
  		& 50-75
 		& 76.2 $\pm$ 2.1
 		& 52.8 $\pm$ 4.1
 		& 7.99 $\pm$ 0.24
  		\\
            &
  		& 75-
 		& 75.2 $\pm$ 0.8
 		& 54.8 $\pm$ 1.5
 		& 8.19 $\pm$ 0.17
  		\\
            &
  		& Average Gap
 		& 0.7
 		& \textbf{0.2}
 		& \textbf{0.67}
  		\\
 		\bottomrule
 	\end{tabular}
 	}
 \label{agegap}
\end{table*}

\section{Clinical Implication and Potential Future Works}
\label{appendix_clinicalimplication}
Our proposed supervised DML and self-supervised DML non-invasively estimate central hemodynamics using the ECG signal. Current non-invasive methods for assessing a number of hemodynamic measurements include clinical analysis using echocardiography \citep{nagueh1997doppler, nagueh2009recommendations, tolia2018validating}. However, this method requires trained personnel to obtain ultrasound images and a specialist to interpret those images, hence it is challenging to use this information in routine practice. Indeed, such information is not readily available in many healthcare settings. On the contrary, our methods will be easily incorporated into clinical workflows to aid in early diagnosis and continuous patient monitoring, and thus contribute to personalized patient management. For instance, the model can be modified to get the ECG signal from one lead to get the limb lead signal from a smartwatch or portable device, and connected to a mobile app for heart failure patient early monitoring. Or, it can be connected to implantable CardioMEMS$^{TM}$ \citep{abraham2011wireless, abraham2016sustained, tolia2018validating} to get the measured hemodynamics to monitor patient status. Our approach would help screen out the patients with abnormal mPCWP and help physicians assess treatment response.

\end{document}